\documentclass[10pt,journal,compsoc]{IEEEtran}
\ifCLASSOPTIONcompsoc
  \usepackage[nocompress]{cite}
\else
  \usepackage{cite}
\fi
\ifCLASSINFOpdf
\else
\fi
\usepackage{url}

\usepackage{algorithm}
\usepackage{algorithmic}
\usepackage{multirow}
\usepackage{latexsym}
\usepackage{subfigure}
\usepackage{graphicx}
\usepackage{amsmath}
\usepackage{amssymb}
\usepackage{hyperref}
\hypersetup{hypertex=true,
	colorlinks=true,
	linkcolor=red,
	anchorcolor=red,
	citecolor=green}
\usepackage{color}
\usepackage{ragged2e}

\usepackage{subfigure}

\usepackage{makecell}
\usepackage{gensymb}

\begin{document}
%
\title{Prototype Optimization with Neural ODE for Few-Shot Learning}
%
%
%
%

\author{Baoquan~Zhang,
        Shanshan~Feng,
        Bingqi~Shan,
        Xutao~Li,
        Yunming~Ye,
        and~
        Yew-Soon Ong,~\IEEEmembership{Fellow,~IEEE}
\IEEEcompsocitemizethanks{\IEEEcompsocthanksitem Baoquan Zhang is with the School of Computer Science and Technology, Harbin Institute of Technology, Shenzhen, Shenzhen 518055, Guangdong, China. E-mail: zhangbaoquan@stu.hit.edu.cn. \protect\\
 Shanshan Feng is with Centre for Frontier Al Research, Institute of High Performance Computing, A*STAR, Singapore. \protect\\
 Bingqi Shan, Xutao Li, and Yunming Ye are with the School of Computer Science and Technology, Harbin Institute of Technology, Shenzhen, 518055, Guangdong, China.\protect\\
Yew-Soon Ong is with the Agency for Science, Technology and Research(A*STAR), Singapore and also with Nanyang technological lniversity, Singapore. \protect\\
\IEEEcompsocthanksitem Corresponding authors are Baoquan Zhang; E-mail: baoquanzhang@hit.edu.cn.\protect\\
\IEEEcompsocthanksitem Code is available at \url{https://github.com/zhangbq-research/MetaNODE}.\protect\\
}
\thanks{Manuscript received August 11, 2021.}
}

%
%

\markboth{Journal of \LaTeX\ Class Files,~Vol.~14, No.~8, August~2015}%
{Shell \MakeLowercase{\textit{et al.}}: Bare Demo of IEEEtran.cls for Computer Society Journals}
%



\IEEEtitleabstractindextext{%
\begin{abstract}
\justifying{Few-Shot Learning (FSL) is a challenging task, which aims to recognize novel classes with few examples. Pre-training based methods effectively tackle the problem by pre-training a feature extractor and then performing class prediction via a cosine classifier with mean-based prototypes. Nevertheless, due to the data scarcity, the mean-based prototypes are usually biased. In this paper, we attempt to diminish the prototype bias by regarding it as a prototype optimization problem. To this end, we propose a novel prototype optimization framework to rectify prototypes, \emph{i.e.}, introducing a meta-optimizer to optimize prototypes. Although the existing meta-optimizers can also be adapted to our framework, they all overlook a crucial gradient bias issue, \emph{i.e.}, the mean-based gradient estimation is also biased on sparse data. To address this issue, in this paper, we regard the gradient and its flow as meta-knowledge and then propose a novel Neural Ordinary Differential Equation (ODE)-based meta-optimizer to optimize prototypes, called MetaNODE. Although MetaNODE has shown superior performance, it suffers from a huge computational burden. To further improve its computation efficiency, we conduct a detailed analysis on MetaNODE and then design an effective and efficient MetaNODE extension version (called E$^2$MetaNODE). It consists of two novel modules: E$^2$GradNet and E$^2$Solver, which aim to estimate accurate gradient flows and solve optimal prototypes in an effective and efficient manner, respectively. Extensive experiments show that 1) our methods achieve superior performance over previous FSL methods and 2) our E$^2$MetaNODE significantly improves computation efficiency meanwhile without performance degradation.}
	
\end{abstract}

\begin{IEEEkeywords}
Few-Shot Learning, Meta-Learning, Image Classification, Prototype Optimization, Meta-Optimizer.
\end{IEEEkeywords}}

\maketitle

\IEEEdisplaynontitleabstractindextext

%
\IEEEpeerreviewmaketitle

\IEEEraisesectionheading{\section{Introduction}\label{sec:introduction}}

%
%
%
%
\IEEEPARstart{W}{ith} abundant annotated data, deep learning techniques have shown very promising performance for many applications, \emph{e.g.}, computer vision \cite{he16} and natural language processing \cite{he16}. However, preparing enough annotated samples is very time-consuming, laborious, or even impractical in some scenarios, \emph{e.g.}, cold-start recommendation \cite{zheng2021}, medical diagnose \cite{PrabhuKRCSA19}, and remote sensing analysis \cite{PrabhuKRCSA19}. Few-shot learning (FSL) has been proposed and received considerable attentions for addressing the issue by mimicking the flexible adaptation ability of human from very few examples. Its main rationale is to learn meta-knowledge from base classes with sufficient labeled samples and then employ the meta-knowledge to perform class prediction for novel classes with few labeled samples \cite{LiLX0019}.

\begin{figure}
	\centering
	\subfigure[Prior Works]{ 
		\label{fig0a} 
		\includegraphics[width=0.47\columnwidth]{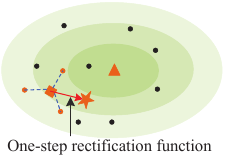}}
	\subfigure[Our Method]{ 
		\label{fig0b} 
		\includegraphics[width=0.47\columnwidth]{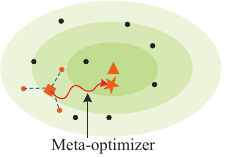}}
	\caption{Pre-training based method estimates prototypes in an average manner, which suffers from a prototype bias issue. Prior works diminish the bias in a one-step manner (a). Our method addresses it in a meta-optimization manner (b). Here, orange and black points denote training and test samples, respectively. Orange square, star, and triangle denotes the mean-based, rectified, and real prototypes, respectively. }
	\label{fig0}
\end{figure}

Previous studies primarily address the FSL problem using the idea of meta-learning, \emph{i.e.}, constructing a large set of few-shot tasks on base classes to learn task agnostic meta-knowledge \cite{FlennerhagRPVYH20}. Recently, Chen et al. \cite{Yinbo20} regard feature representation as meta-knowledge, and propose a simple pre-training method, which delivers more promising performance. In the method, they first pre-train a feature extractor on all base classes, and then perform novel class prediction via mean-based prototypes. However, this method suffers from a {\bf prototype bias issue}, \emph{i.e.}, the discrepancy between calculated mean and real prototypes. As illustrated in Figure~\ref{fig0}, the mean-based prototype (orange square) is usually far away from real prototype (triangle). This kind of prototype bias is caused by the fact that scarce labeled samples cannot provide a reliable mean estimation for the prototypes \cite{YaohuiWang_pr}. To address the drawback, , as shown in Figure~\ref{fig0a}, some prior works attempt to learn a one-step prototype rectification function from a large set of few-shot tasks \cite{YaohuiWang_pr, XueW20, zhang2021}, and then attempt to rectify the prototype bias in a one-step manner. However, due to the 
complexity of prototype bias, characterizing the bias with a one-step rectification function is too coarse to obtain accurate prototypes (see Table~\ref{table3}).

In this paper, we propose a novel meta-learning based prototype optimization framework to rectify the prototype bias. In the framework, instead of using the one-step rectification manner, we consider the bias reduction as a prototype optimization problem and attempt to diminish the prototype bias with an optimization-based meta-learning method (called meta-optimizer). The idea behind such design is learning a novel Gradient Descent Algorithm (GDA) on base classes and then utilizing it to polish the prototypes via a few gradient update steps for novel classes. Specifically, we first pre-train a classifier on all base classes to obtain a good feature extractor. Then, given a few-shot task, as shown in Figure~\ref{fig0b}, we average the extracted features of all labeled samples as the initial prototype for each class. As a sequel, these prototypes will be further optimized to reduce the prototype bias through a meta-optimizer. Finally, we perform the class prediction via a nearest neighbor classifier.  

The workhorse of our framework is the meta-optimizer, in which the mean-based prototypes will be further polished through GDA. Even though the existing meta-optimizer such as ALFA \cite{BaikCCKL20} and MetaLSTM \cite{RaviL17} can also be utilized for this purpose, 
they all suffer from a common drawback, called {\bf gradient bias issue}, \emph{i.e.}, their gradient estimation is inaccurate on sparse data. The issue appears because all the existing meta-optimizers carefully model the hyperparameters (\emph{e.g.}, initialization \cite{RaghuRBV20} and regularization parameters \cite{BaikCCKL20, FlennerhagRPVYH20}) in GDA as meta-knowledge, but roughly estimate the gradient in an average manner with very few labeled samples, which is usually inaccurate. Given that the gradient estimation is inaccurate, more excellent hyperparameters are not meaningful and cannot lead to a stable and reliable prototype optimization. Thus, how design a new meta-optimizer with accurate gradient estimation is the key challenge of our prototype optimization framework. 

To address the challenge, in our conference version \cite{zhang2022metanode}, we treat the gradient and its flow in GDA as meta-knowledge, and propose a  Neural Ordinary Differential Equation (ODE)-based meta-optimizer (called MetaNODE) to model prototype optimization as continuous-time dynamics specified by a Neural ODE. The idea is inspired by the fact that the GDA formula is indeed an Euler-based discrete instantiation of an ODE \cite{bu2020}, and the ODE will turn into a Neural ODE \cite{ChenRBD18} when we treat its gradient flow as meta-knowledge. In particular, we first design a gradient flow inference network (GradNet), which aims to estimate a accurate continuous-time gradient flow for prototype dynamics. Then, given an initial prototype (\emph{i.e.}, the mean-based prototype), the optimal prototype can be obtained by solving the Neural ODE in a RK4-based ODE solver. The advantage of such design is the process of prototype rectification can be characterized in a continuous manner, thereby producing more accurate prototypes. 

Although the MetaNODE proposed in conference version \cite{zhang2022metanode} has achieved superior performance, in current extension version, we find that the MetaNODE suffers from a computation burden issue on running time. To figure out the key reasons, in this extension version, we conduct a detailed analysis about computation complexity of our MetaNODE (see Table~\ref{table222}). We find that such computation burden is introduced by our GradNet (due to its complex calculation of multi-modules ensemble and multi-head attention) and RK4-based ODE solver (due to its high-order integral). To reduce the computation complexity, we further simplify our GradNet and RK4-based ODE solver and then develop an efficient and effective extension version of our MetaNODE (called E$^2$MetaNODE). Specifically,  \textbf{1) in the GradNet aspect}, we theoretically analyze its working rationale from the perspective of gradient flows of prototype classifier and then simplify the GradNet into a more simple, efficient and effective network (called E$^2$GradNet). E$^2$GradNet effectively improves computation efficiency meanwhile without performance degradation. \textbf{2) In the ODE Solver aspect}, we design an efficient and effective meta-learning based ODE Solver (called E$^2$Solver) by treating the integral error of ODE solver as meta-knowledge. It aims to learn to estimate the integral error for more accurately solving the Neural ODE. The E$^2$Solver effectively reduces the integral error meanwhile with comparable inference time with a low-order Euler-based ODE Solver.

Our main contributions can be summarized as follows:

\begin{itemize}
	\item We propose a new perspective to rectify prototypes for FSL, by regarding the bias reduction problem as a prototype optimization problem, and present a novel meta-learning based prototype optimization framework to improve FSL performance. 
	\item We identify a crucial issue of existing meta-optimizers
	, \emph{i.e.}, gradient bias issue. To address the issue, we propose a novel Neural ODE-based meta-optimizer (MetaNODE) by modeling the process of prototype optimization as continuous-time dynamics specified by a Neural ODE. Our MetaNODE optimizer can effectively alleviate the gradient bias issue and leads to more accurate prototypes. 
	\item We further develop an MetaNODE extension version (called E$^2$MetaNODE) for efficient computation. First, we theoretically analyze the rationale of our GradNet and then simplify its design in a more effective and efficient network (called E$^2$GradNet), which improves computation efficiency meanwhile without performance degradation. Second, we treat the integral error of ODE solver as meta-knowledge and then design a meta-learning based solver (called E$^2$Solver), which obtains more accuracy prototype estimation meanwhile with comparable inference time with Euler-based ODE solvers.
	\item We conduct comprehensive experiments on both transductive and inductive FSL settings, which demonstrate the effectiveness and efficiency of our MetaNODE and E$^2$MetaNODE.
\end{itemize}

This paper is an extension to our conference version in \cite{zhang2022metanode}. Compared to the conference paper, this version additionally presents (\romannumeral1) a theoretic analysis on the gradient flow of prototype classifier, which clarifies why our GradNet works; (\romannumeral2) a more effective and efficient GradNet (i.e., E$^2$GradNet), which reduces the computation cost significantly meanwhile without performance degrade; (\romannumeral3) a meta-learning based ODE solver (i.e., E$^2$Solver), which obtains more accuracy prototype estimation meanwhile with comparable inference time with Euler solver. The above E$^2$GradNet and E$^2$Solver form our MetaNODE extension version (i.e., E$^2$MetaNODE); (\romannumeral4) more performance evaluation on CIFAR-FS and FC100, statistical analysis, ablation results, and visualization on miniImagenet, tieredImageNet, CUB-200-2011, and comparisons with more state-of-the-art methods in both transductive and inductive FSL settings.

The rest of this work is organized as follows: In Section~\ref{section_2}, we briefly review related works on few-shot learning including inductive FSL and transductive FSL, and Neural ODE. Section~\ref{section_3} describes our meta-learning-based prototype optimization framework. Section~\ref{section_3_3} introduce our MetaNODE meta-optimizer and its gradient flow inference network in details. Section~\ref{section_4} presents the effective and efficient extension version (i.e., E$^2$MetaNODE) including the computation efficiency analysis, GradNet theoretic analysis and its two key effective and efficient extension components, \emph{i.e.}, $E^2$GradNet and $E^2$Solver. Section~\ref{section_5} presents the results. Finally, the conclusion is summarized in Section~\ref{section_6}.

\section{Related Work}
\label{section_2}


\subsection{Few-Shot Learning}
FSL is a challenging task, aiming to recognize novel classes with few labeled samples. According to the test setting, FSL can be divided into two groups, \emph{i.e.}, inductive FSL and transductive FSL \cite{hospedales2021meta}. The former assumes that information from test data cannot be utilized when classifying the novel class samples while the latter considers that all the test data can be accessed to make novel class prediction.

\subsubsection{Inductive Few-Shot Learning}
In earlier studies, most methods mainly focus on inductive FSL setting, which can be roughly grouped into three categories.  1) Metric-based approaches. This line of works focuses on learning a task-agnostic metric space and then predicting novel classes by a nearest-centroid classifier with Euclidean or cosine distance such as \cite{NguyenLWKRJ20, SnellSZ17, li2019finding, yang2021bridging, zhang2022deepemd, cheng2023frequency, zhang2023deta}. For example, \cite{li2019finding} proposed a category traversal module to find task-relevant features, aiming to classify each sample in a low-dimension and compact metric space. In \cite{zhang2022deepemd}, zhang et al. employ the Earth Mover's Distance (EMD) as a metric to compute a structural distance between dense feature representations for FSL. 2) Optimization-based approaches. The key idea is to model an optimization algorithm (\emph{i.e.}, GDA) over few labeled samples within a meta-learning framework \cite{Johannes21, RaghuRBV20, baik2023learning, kang2023meta, sun2020meta, gao2022curvature, du2023emo}, which is known as meta-optimizer, such as MetaLSTM \cite{RaviL17} and ALFA \cite{BaikCCKL20}. For example, MAML \cite{FinnAL17}, MetaLSTM \cite{RaviL17}, adn ALFA \cite{BaikCCKL20} introduced meta-learners to learn to design initialization, update rule, and weight decay of optimization algorithm, respectively. In \cite{gao2022curvature}, Gao et al. propose  a curvature-adaptive meta-learning method for fast adaptation to manifold data by producing suitable curvature. In \cite{du2023emo}, Du et al. propose an external memory to store the gradient history and then leverage it to learn more accurate gradient for meta-optimizer such that the optimal model parameters can be more accurately and quickly updated. 3) Pre-training based approaches. This type of works mainly utilizes a two-phases training manner to quickly adapt to novel tasks, \emph{i.e.}, pre-training and fine-tuning phases, such as \cite{LiuCLL0LH20, ShenLQSC21, Yinbo20, Zhiuncertainty21, LiFG21, zhang2023prototype, zhang2024metadiff, du2023protodiff}. For instance, Chen et al. \cite{Yinbo20} proposed to train a feature extractor from the entire base classes and then classify novel classes via a nearest neighbor classifier with cosine distance for few-shot tasks, which suffers from the prototype bias issue. In \cite{du2023protodiff}, Du et al. proposed a task-guided diffusion model for prototype learning, which aims to remove the noise contained in class prototypes in a diffusion manner.

In this paper: 1) we also focus on prototype bias issue but attempt to address it in a continuous-time prototype optimization manner; 2) we design a novel continuous-time meta-optimizer by regarding the gradient itself and ODE solver as meta-knowledge for prototype optimization.

\subsubsection{Transductive Few-Shot Learning}
Recently, several studies explored transductive FSL setting and showed superior performance on FSL, which can be divided into two categories. 1) Graph-based approaches. This group of methods attempts to propagate the labels from labeled samples to unlabeled samples by constructing an instance graph for each FSL task, such as \cite{Chen_2021_CVPR, YangLZZZL20, Tang_2021_CVPR}. For instance, Chen et al. \cite{LiuLPKYHY19} proposed a transductive propagation network to learn to propagate labels for FSL. In \cite{YangLZZZL20}, a distribution propagation graph network was devised for FSL, which explores distribution-level relations for label propagations. Chen et al. \cite{RodriguezLDL20} further extended \cite{LiuLPKYHY19} by exploring embedding propagation, to obtain a smooth manifold for label propagation. 
2) Pre-training based approaches. This kind of studies still focuses on the two-stages training paradigms. Different from inductive FSL methods, these methods further explore the unlabeled samples to train a better classifier \cite{Malik20, HuMXSOLD20, WangXLZF20, ZikoDGA20, zhu2023transductive, tian2023prototypes} or construct more reliable prototypes \cite{XueW20, zhang2021}. For example, Ziko et al. \cite{ZikoDGA20} proposed a Laplacian-based regularization from the perspective of training a new classifier to encourage label consistency of nearby unlabeled samples. Liu et al. \cite{YaohuiWang_pr} proposed a label propagation and a feature shifting strategy for more reliable prototype estimation. 

In this paper, we also target at obtaining reliable prototypes for FSL. Different from these existing methods, we regard it as a prototype optimization problem and propose a new Neural ODE-based meta-optimizer for optimizing prototypes. Besides, our method is more flexible, which can be applied to both transductive and inductive FSL settings. 


\begin{figure*}[t]
	\centering
	\includegraphics[width=1.0\textwidth]{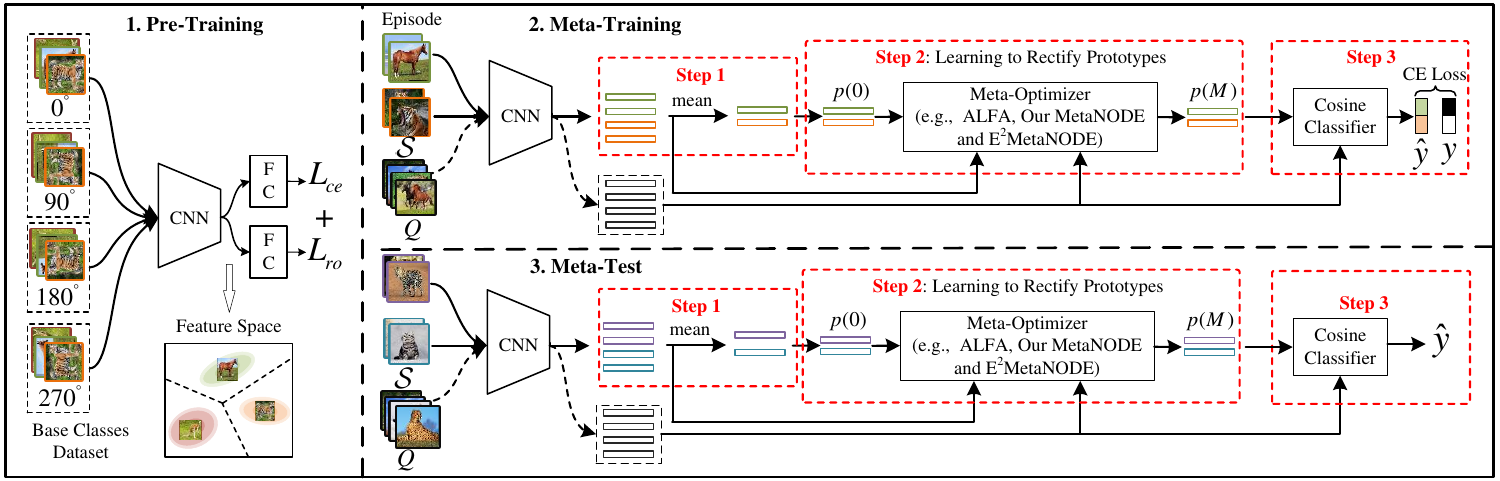} 
	\caption{The meta-learning based prototype optimization framework, which consists pre-training, meta-training, and meta-test phases. Among them, the pre-training phase aims to learn a feature
    extractor to obtain a good image representation for each image. Then, a meta-optimizer is introduced to learn to rectify prototypes for each class during meta-training phase. Finally, the meta-test is employed to perform few-shot class prediction.}
  \vspace{-15pt}
	\label{fig2}
\end{figure*}


\subsection{Neural ODE}
Neural ODE, proposed by \cite{ChenRBD18}, is a continuous-time model, aiming to capture the evolution process of a state by representing its gradient flow with a neural network. Recently, it has been successfully used in various domains, such as irregular time series prediction \cite{RubanovaCD19}, knowledge graph forecasting \cite{abs210105151}, MRI image reconstruction \cite{ChenCS20}, image dehazing \cite{ShenL0X020}, diffusion generation \cite{yang2023diffusion, ho2020denoising}. For example, in \cite{RubanovaCD19}, Shen et al. propose a novel recurrent neural networks with continuous-time hidden dynamics defined by ordinary differential equations (ODEs) for irregularly-sampled time serie prediction. In \cite{ho2020denoising}, Ho et al. regard the image generation as a stochastic ODE dynamic from a Gaussian distribution to target image distribution. However, to our best knowledge, there is few previous works to explore it for FSL. 
In this paper, we propose a Neural ODE-based meta-optimizer to polish prototypes. Its advantage is that prototype dynamics can be captured in a continuous manner, which produces more accurate prototypes for FSL. 


\section{Problem Definition and Framework}
\label{section_3}


\subsection{Problem Definition}
\label{section_3_1}
For a $N$-way $K$-shot problem, two datasets are given: a base class dataset $\mathcal{D}_{base}$ and a novel class dataset $\mathcal{D}_{novel}$. The base class dataset $\mathcal{D}_{base}=\{(x_i, y_i)\}_{i=0}^{B}$ is made up of abundant labeled samples, where each sample $x_i$ is labeled with a base class $y_i \in \mathcal{C}_{base}$ ($\mathcal{C}_{base}$ denotes the set of base classes). The novel class dataset consists of two subsets: a training set $\mathcal{S}$ with few labeled samples (called support set) and a test set $\mathcal{Q}$ consisting of unlabeled samples (called query set). Here, the support set $\mathcal{S}$ is composed of $N$ classes sampled from the set of novel class $\mathcal{C}_{novel}$, and each class only contains $K$ labeled samples. Note that the base class set and novel class set are disjoint, \emph{i.e.}, $\mathcal{C}_{base} \cap \mathcal{C}_{novel} = \emptyset$. 

For transductive FSL, we regard all query samples $x \in \mathcal{Q}$ as unlabeled sample set $\mathcal{Q}'$. Our goal is to learn a classifier for query set $\mathcal{Q}$ by leveraging unlabeled sample set $\mathcal{Q}'$, support set $\mathcal{S}$, and base class dataset $\mathcal{D}_{base}$. However, for inductive FSL, the classifier is obtained only by leveraging $\mathcal{S}$ and  $\mathcal{D}_{base}$. In following subsections, we focus on transductive FSL to introduce our method, and how to adapt to inductive FSL will be explained at Section~\ref{section_3_3}. 

\subsection{Prototype Optimization Framework}
\label{section3_2}
In this paper, we focus on addressing the prototype bias issue appearing in the pre-training FSL method \cite{Yinbo20}. Different from existing one-step rectification methods \cite{YaohuiWang_pr, XueW20}, we regard the bias reduction as a prototype optimization problem and present a novel meta-learning based prototype optimization framework to rectify prototypes. Our idea is introducing a prototype meta-optimizer to learn to diminish the prototype bias. As shown in Figure~\ref{fig2}, the framework consists of three phases, \emph{i.e.}, pre-training, meta-training, and meta-test phases. Next, we detail on them, respectively.  

\noindent {\bf Pre-Training.} Following \cite{RodriguezLDL20}, we first pretrain a feature extractor $f_{\theta_f}()$ with parameters $\theta_f$ by minimizing both a classification loss $L_{ce}$ and an auxiliary rotation loss $L_{ro}$ on all base classes. This aims to obtain a good image representation. Then, the feature extractor is frozen. 

\noindent {\bf Meta-Training.} 
Upon the feature extractor $f_{\theta_f}()$, we introduce a meta-optimizer $g_{\theta_g}()$ to learn to rectify prototypes in an episodic training paradigm \cite{VinyalsBLKW16}. The idea behind such design is that learning task-agnostic meta-knowledge about prototype rectification from base classes and then applying this meta-knowledge to novel classes to obtain more reliable prototypes for FSL. The main details of the meta-optimizer will be elaborated in Section~\ref{section_3_3} and \ref{section_4}. Here, we first introduce the workflow depicted in Figure~\ref{fig2}.  

As shown in Figure~\ref{fig2}, following the episodic training paradigm \cite{VinyalsBLKW16}, we first mimic the test setting and construct a number of $N$-way $K$-shot tasks (called episodes) from base class dataset $\mathcal{D}_{base}$. For each episode, we randomly sample $N$ classes from base classes $\mathcal{C}_{base}$, $K$ images per class as support set $S$, and $M$ images per class as query set $Q$. Then, we train the  above meta-optimizer $g_{\theta_g}()$ to polish prototypes by minimizing the negative log-likelihood estimation on the query set $\mathcal{Q}$. That is,
\begin{equation}
	\begin{aligned}
		\min\limits_{\theta_g}  \mathbb{E}_{(\mathcal{S},\mathcal{Q}) \in \mathbb{T}} \sum_{(x_i,y_i) \in \mathcal{Q}} -log(P(y_i|x_i, \mathcal{S}, \mathcal{Q}', \theta_g)),
	\end{aligned}
	\label{eq5_0}
\end{equation}
where $\mathbb{T}$ is the set of constructed $N$-way $K$-shot tasks and $\theta_g$ denotes the parameters of the meta-optimizer $g_{\theta_g}()$. Next we introduce how to calculate the class probability $P(y_i|x_i, \mathcal{S}, \mathcal{Q}', \theta_g)$, including the following three steps: 

{\bf Step 1.} We first leverage the feature extractor $f_{\theta_f}()$ to represent each image. Then we compute the mean-based prototype of each class $k$ as initial prototype $p_k(0)$ at $t=0$:
\begin{equation}
	p_k(0) = \frac{1}{|\mathcal{S}_k|} \sum_{(x_i, y_i) \in \mathcal{S}_k} f_{\theta_f}(x_i),
	\label{eq5}
\end{equation}
where $\mathcal{S}_k$ is the support set extracted from class $k$ and $t$ denotes the iteration step (when existing meta-optimizers are employed) or the continuous time (when our MetaNODE and E$^2$MetaNODE described in Section~\ref{section_3_3} and Section~\ref{section_4} are utilized). For clarity, we denote the prototype set $\{p_k(t)\}_{k=0}^{N-1}$ as the prototypes $p(t)$ of classifiers at iteration step/time $t$, \emph{i.e.}, $p(t) = \{p_k(t)\}_{k=0}^{N-1}$.

{\bf Step 2:} Unfortunately, the initial prototypes $p(0)$ are biased since only few support samples are available. To eliminate the bias, we view the prototype rectification as an optimization process. Then, given the initial prototypes $p(0)$, the optimal prototypes $p(M)$ can be obtained by leveraging the meta-optimizer $g_{\theta_g}()$ to optimize prototypes. That is, 
\begin{equation}
	p(M) = \Psi(g_{\theta_{g}}(), p(0), \mathcal{S}, \mathcal{Q}', t=M),
	\label{eq6}
\end{equation}where $M$ is the total iteration number/integral time and $\Psi()$ denotes the process of prototype optimization. Please refer to Section~\ref{section_3_3} and Section~\ref{section_4} for the details of the meta-optimizer optimization process. 


{\bf Step 3:} Finally, we regard the optimal prototypes $p(M)$ as the final prototypes. Then, we evaluate the class probability that each sample $x_i \in \mathcal{Q}$ belongs to class $k$ by computing the cosine similarity between $x_i$ and $p(M)$. That is,
\begin{equation}
	P(y=k|x_i, \mathcal{S}, \mathcal{Q}', \theta_g) = \frac{e^{\gamma \cdot <f_{\theta_f}(x_i), p_{k}(M)>}}{\sum_c e^{\gamma \cdot <f_{\theta_f}(x_i), p_{c}(M)>}},
	\label{eq7}
\end{equation}
where $<\cdot>$ denotes the cosine similarity, and $\gamma$ is a scale parameter. Following \cite{ChenLKWH19}, $\gamma=10$ is used. 


\noindent {\bf Meta-Test.} Its workflow is similar to the meta-training phase. The difference is that we remove the meta-optimizer training step defined in Eq.~\ref{eq5_0} and directly perform few-shot classification for novel classes by following Eqs.~\ref{eq5} $\sim$ \ref{eq7}.

\section{Meta-Optimizer: MetaNODE}
\label{section_3_3}
In the prototype framework described in Section~\ref{section3_2}, the key challenge is how to design a meta-optimizer to polish prototypes. In this subsection, we first discuss the limitation of existing meta-optimizers to polish prototypes. Then, a novel meta-optimizer, \emph{i.e.}, MetaNODE, is presented. 

\subsection{\textbf{Analyses on Existing Meta-Optimizers}} In the above framework, several existing meta-optimizers can be utilized to diminish prototype bias by setting the prototypes as their variables to be updated, such as MetaLSTM \cite{RaviL17} and ALFA \cite{BaikCCKL20}. However, we find that they suffer from a new drawback, \emph{i.e.}, gradient bias issue. Here, we take ALFA as an example to illustrate the issue. Formally, for each few-shot task, let $L(p(t))$ be its differentiable loss function (Following \cite{BaikCCKL20}, the cross-entropy loss is used in this paper) with prototype $p(t)$, and $\nabla L(p(t))$ be its prototype gradient, 
during performing prototype optimization. Then, the ALFA can be expressed as the following $M$-steps iterations, given the initial (\emph{i.e.}, mean-based) prototypes $p(0)$. That is,
\begin{equation}
	p(t+1) = p(t) - \eta (\nabla L(p(t))+\omega p(t)),
	\label{eq1}
\end{equation}
where $t$ is the iteration step (\emph{i.e.}, $t=0, 1, ..., M-1$), $\eta$ is a learning rate, and $\omega$ denotes a weight of $\ell_2$ regularization term infered by the meta-optimizer $g_{\theta_g}()$. Its goal is to improve fast adaptation with few examples by learning task-specific $\ell_2$ regularization term. Though ALFA is effective (see Table~\ref{table5}), we find that it computes the gradient $\nabla L(p(t))$ in an average manner over few labeled samples $(x_i, y_i) \in \mathcal{S}$:
\begin{equation}
	\nabla L(p(t)) = \frac{1}{|\mathcal{S}|} \sum_{(x_i,y_i) \in \mathcal{S}} \nabla L_{(x_i,y_i)}(p(t)),
	\label{eq2}
\end{equation}
where $|\cdot|$ denotes the size of a set and $\nabla L_{(x_i,y_i)}(p(t))$ is the gradient of the sample $(x_i,y_i) \in \mathcal{S}$. Such estimation is inaccurate, because the number of available labeled (support) samples (\emph{e.g.}, $K$=1 or 5) is far less than the expected amount. As a result, the optimization performance of existing methods is limited. This is exactly the gradient bias issue mentioned in the section of introduction. Note that the unlabeled samples $x \in \mathcal{Q}'$ are not used in these methods because their gradients cannot be computed without labels. 

\begin{figure}[t]
	\centering
	\includegraphics[width=0.95\columnwidth]{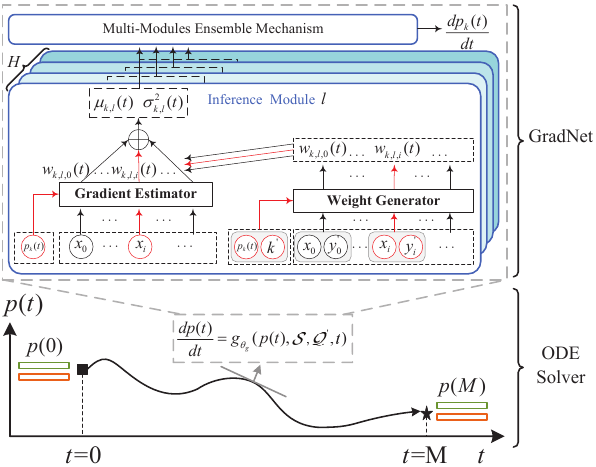} 
	\caption{Illustration of our MetaNODE, which consists of a GradNet and an ODE Solver. The former aims to infer the continuous-time prototype gradient flow, and the latter accounts for solving the Neural ODE to obtain the polished prototypes. The red lines represent the computation flow of the sample $x_i \in \mathcal{S} \cup \mathcal{Q}'$. Note that the gradient estimator and the weight generator are shared with all samples $x_i$ in the set of  $\mathcal{S} \cup \mathcal{Q}'$.}
  \vspace{-15pt}
	\label{fig3}
\end{figure}

\subsection{\textbf{MetaNODE} }
Recent studies \cite{bu2020} found that the iteration of Gradient Descent Algorithm (GDA) can be viewed as an Euler discretization of an ordinary differential equation (ODE), i.e.,
\begin{equation}
	\frac{\mathrm{d} p(t)}{\mathrm{d} t} = - \nabla L(p(t)),
	\label{eq_7777}
\end{equation}
where $t$ is a continuous variable (\emph{i.e.}, time) and $\frac{\mathrm{d} p(t)}{\mathrm{d} t}$ denotes a continuous-time gradient flow of prototypes $p(t)$; and $L(p(t))$ denotes a cross-entropy loss over over prototype $P(t)$ on the support set $\mathcal{S}$ for each few-shot task.

Inspired by this fact, to more finely rectify the prototypes, we propose to characterize the prototype dynamics by an ODE and consider the prototype rectification problem as the ODE initial value problem, where the initial and final status values correspond to the mean-based and optimal prototypes, respectively. To address the gradient bias issue appeared in Eq. \ref{eq2}, we view the prototype $p(t)$, support set $\mathcal{S}$, unlabeled sample set $\mathcal{Q}'$, and time $t$ as inputs and then employ a neural network (\emph{i.e.} the meta-learner $g_{\theta_g}$()) to directly estimate the continuous gradient flow $\frac{\mathrm{d} p(t)}{\mathrm{d} t}$. The advantage of such design is that although the number of support examples does not change, the meta-knowledge using for gradient estimation from lots of base class tasks can be effectively leveraged to enhance the gradient estimation for few-shot novel classes, such that more accurate gradient estimation can be obtained than simple mean-based gradient estimation. Then, the ODE turns into a Neural ODE, \emph{i.e.}, $\frac{\mathrm{d} p(t)}{\mathrm{d} t} = g_{\theta_g}(p(t), \mathcal{S}, \mathcal{Q}', t)$. 

Based on this notion, we design a novel Neural ODE-based meta-optimizer (MetaNODE). Its advantage is that the prototype rectification dynamics can be captured in a continuous manner, thereby more finely diminishing the prototype bias. As shown in Figure~\ref{fig3}, the MetaNODE consists of a Gradient Flow Inference Network (GradNet) and an ODE solver. The GradNet is regarded as the meta-learner $g_{\theta_g}()$, aiming to infer the continuous gradient flow $\frac{\mathrm{d} p(t)}{\mathrm{d} t}$ (see next subsection for its details). Based on the GradNet $g_{\theta_g}()$ and initial prototypes $p(0)$, the optimal prototypes $p(M)$ can be obtained by evaluating Neural ODE at the last time point $t=M$, \emph{i.e.}, $p(M) = p(0) + \int_{t=0}^M g_{\theta_g}(p(t), \mathcal{S}, \mathcal{Q}', t)$, where the integral term is calculated by ODE solvers. That is,
\begin{equation}
	p(M) = ODESolver(g_{\theta_g}(), p(0), \mathcal{S}, \mathcal{Q}', t=M).
	\label{eq3}
\end{equation}
Following \cite{ChenRBD18}, we use RK4 method \cite{Alexander90} as our ODE solver because it has relatively low integration error.  

\begin{table*}
	\caption{Test time comparsion on miniImagenet. Here, ``E'' and ``A'' denotes the component of ensemble and attention, respectively.}\smallskip
	\centering
	\smallskip\scalebox
	{0.98}{
		\smallskip\begin{tabular}{c|c|c|c|c|c|c|c}
			\hline
			\multirow{2}{*}{Setting} & \multirow{2}{*}{ClassifierBaseline \cite{Yinbo20}} & \multirow{2}{*}{ALFA \cite{BaikCCKL20}} & \multirow{2}{*}{CloserLook \cite{ChenLKWH19}} & \multicolumn{2}{c|}{\multirow{1}{*}{Our MetaNODE}}  & \multicolumn{2}{c}{\multirow{1}{*}{Our MetaNODE (w/o E and A)}} \\
			\cline{5-8}
			& & & & Euler Solver & RK4 Solver & Euler Solver & RK4 Solver \\
			\hline
			5-way 1-shot & 0.046 s & 0.081 s & 0.133 s & 0.344 s & 1.290 s &  0.116 s & 0.194 s \\
			5-way 5-shot & 0.059 s & 0.085 s & 0.359 s & 0.452 s & 1.300 s &  0.132 s &  0.210 s \\
			\hline
	\end{tabular}}
	\vspace{-15pt}
	\label{table222}
\end{table*}

\subsection{\textbf{Gradient Flow Inference Network}}
In this section, we introduce how the GradNet $g_{\theta_g}()$ employed in MetaNODE is designed. Intuitively, different classes have distinct prototype dynamics. For many classes like animals and plants, the differences of their prototype dynamics may be quite large. To model the class diversities, as shown in Figure~\ref{fig3}, instead of performing a single inference module, we design multiple inference modules with the same structure to estimate the prototype gradient $\frac{\mathrm{d} p(t)}{\mathrm{d} t}$. Here, the inference module consists of a gradient estimator and a weight aggregator. The former aims to predict the contributed gradient of each sample $x_i \in \mathcal{S} \cup \mathcal{Q}'$ for prototypes $p(t)$. The latter accounts for evaluating the importance of each sample and then combining their gradient estimations in a weighted mean manner. For clarity, we take inference module $l$ and class $k$ as an example to detail them. 


\noindent {\bf Gradient Estimator.} As we adopt the cosine-based classifier, the prototype is expected to approach the angle center of each class. To eliminate the impact of vector norm, we first transform the features $f_{\theta_f}(x_i)$ of each sample $x_i$ to an appropriate scale by a scale layer $g_{\theta_{gs}^{l}}()$ with parameters $\theta_{gs}^{l}$. Then, the gradient $d_{k, l, i}(t)$ is estimated by computing the difference vector between it and prototype $p_k(t)$, i.e.,
\begin{equation} 
	\begin{aligned}
		d_{k, l, i}(t) = g_{\theta_{gs}^{l}}(f_{\theta_f}(x_i)\|p_k(t)) \otimes f_{\theta_f}(x_i) - p_k(t),
	\end{aligned}
	\label{eq12}
\end{equation}
where $\|$ is a concatenation operation of two vectors and $\otimes$ denotes an element-wise product operation.

\noindent {\bf Weight Generator.} Intuitively, different samples make varying contributions to the gradient prediction of prototype $p_k(t)$. To this end, we design a weight generator to predict their weights. Specially, for each sample $x_i \in \mathcal{S} \cup \mathcal{Q}'$, we combine the prototype $p_k(t)$ and the sample $x_i$ as a new feature. Then, the weight generating process involves a simple feed-forward mapping of the new features by an embedding layer $g_{\theta_{ge}^{l}}()$, followed by a relation layer $g_{\theta_{gr}^{l}}()$ with a multi-head based attention mechanism \cite{VaswaniSPUJGKP17} and an output layer $g_{\theta_{go}^{l}}()$. Here, the relation layer aims to obtain a robust representation by exploring the pair-wise relationship between all samples and the output layer evaluates the contributed weight $w_{k,l,i}$. The above weight generating process can be summarized as Eq.~\ref{eq13}. That is,
\begin{small}
	\begin{equation} 
		\begin{aligned}
			h_{k,l,i}(t)=g_{\theta_{ge}^{l}}(k'\|p_k(t)\|&y'_i\|f_{\theta_f}(x_i)\|p_k(t) \otimes f_{\theta_f}(x_i)),
			\\h_{k,l,i}'(t) =  g_{\theta_{gr}^{l}}&(\{h_{k,l,i}(t)\}_{i=0}^{|\mathcal{S} \cup \mathcal{Q}|-1}),
			\\w_{k,l,i}(t) = &g_{\theta_{go}^{l}}(h_{k,l,i}'(t)),
		\end{aligned}
		\label{eq13}
	\end{equation}
\end{small}where $\theta_{ge}^{l}$, $\theta_{gr}^{l}$, and $\theta_{go}^{l}$ denote model parameters; $k'$ and $y'_i$ denotes the one-hot label of prototype $p_k(t)$ and sample $x_i$, respectively. We replace the one-hot label of each unlabeled sample $x_i \in \mathcal{Q}'$ in a $N$-dim vector with value of $1/N$.

Finally, the gradient $\mu_{k,l}(t)$ and its estimation variance $\sigma^{2}_{k,l}(t)$ can be obtained in a weighted mean manner:
\begin{footnotesize}
	\begin{equation} 
		\begin{aligned}
			\mu_{k,l}(t)=&\sum\nolimits_{i} w_{k,l,i}(t) \otimes d_{k,l,i}(t), \\ \sigma_{k,l}^2(t)=\sum\nolimits_{i} &w_{k,l,i}(t) \otimes (d_{k,l,i}(t) -\ \mu_{k,l}(t))^2.
		\end{aligned}
		\label{eq17}
	\end{equation}
\end{footnotesize}

\noindent {\bf Multi-Modules Ensemble Mechanism.} We have obtained multiple gradient estimations for prototype $p_k(t)$, \emph{i.e.}, $\{\mu_{k,l}(t)\}_{l=0}^{H-1}$, where $H$ denotes the number of inference modules. 
Intuitively, the variance $\{\sigma_{k,l}^{2}(t)\}_{l=0}^{H-1}$ reflects the inconsistency of gradients contributed by all samples, \emph{i.e.}, the larger variance implies greater uncertainty. Hence, to obtain a more reliable prototype gradient $\frac{\mathrm{d} p_k(t)}{\mathrm{d} t}$, we regard the variances as weights to combine these gradients $\mu_{k,l}(t)$: 
\begin{scriptsize}
	\begin{equation} 
		\begin{aligned}
			\frac{\mathrm{d} p_k(t)}{\mathrm{d} t}=\beta\left[\sum_{l=0}^{H-1} (\sigma_{k,l}^{2}(t))^{-1}\right]^{-1}\left[\sum_{l=0}^{H-1} (\sigma_{k,l}^{2}(t))^{-1} \mu_{k,l}(t)\right],
		\end{aligned}
		\label{eq19}
	\end{equation}
\end{scriptsize}where we employ a term of exponential decay with time $t$, \emph{i.e.}, $\beta = \beta_0 \xi^{\frac{t}{M}}$ ($\beta_0$ and $\xi$ are hyperparameters, which are all set to 0.1 empirically), to improve the model stability. 

\subsection{Adaption to Inductive FSL Setting}
\label{sec_4}
The above MetaNODE-based framework focuses on transductive FSL, which can also be easily adapted to inductive FSL. Its workflow is similar to the process described in Sections~\ref{section3_2} and \ref{section_3_3}. The only difference is the unlabeled sample set $\mathcal{Q}'$ is removed in the GradNet of Sections~\ref{section_3_3}, and only using support set $\mathcal{S}$ to estimate gradients.

\section{Meta-Optimizer: E$^2$MetaNODE}
\label{section_4}
\begin{figure}[t]
	\centering
	\includegraphics[width=0.95\columnwidth]{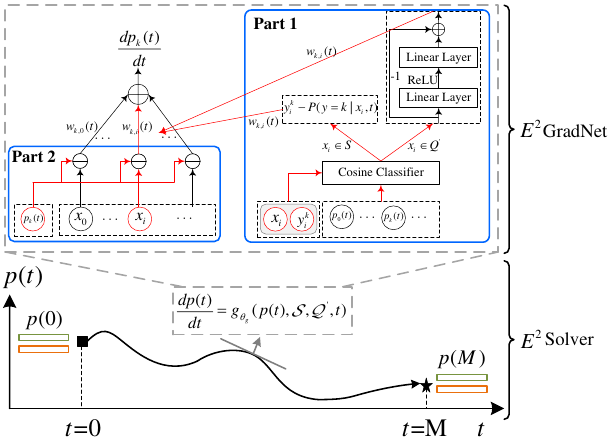} 
	\caption{Illustration of MetaNODE extension version (i.e., E$^2$MetaNODE), which improves its two key components, i.e., GradNet and ODE solver. For GradNet, we simply its design from theoretical perspective to an more simple effective and efficient network (i.e., E2GradNet). For ODE solver, we propose a meta-learning-based ODE solver (i.e., E2Solver) for more accurate and efficient prototype optimization. }
    \vspace{-15pt}
	\label{fig311}
\end{figure}

Till now, we have introduced all details of our prototype
optimization framework and our MetaNODE meta-optimizer. Its advantage is that the prototype bias can be characterized in a continous-time manner such that a more accurate prototype can be obtained for FSL. However, considering the computational complexity of our meta optimizer (i.e., MetaNODE), in this section, we first analyze the theoretical complexity and computation cost of our MetaNODE. Then, we analyze the reason, i.e., why our gradient flow inference nework was designed this way. Finally, based on the above analysis, as shown in Figure~\ref{fig311}, we improve the two key components (i.e., GradNet and ODE solver) of our meta optimizer (i.e., MetaNODE). For GradNet, we simply its design from theoretical perspective to an more simple, effective and efficient network (i.e., E$^2$GradNet). For ODE solver, we propose a meta-learning-based ODE solver (i.e., E$^2$Solver) for more accurate and efficient prototype optimization. The above two components (i.e., E$^2$GradNet and E$^2$Solver) form our efficient and effective meta optimizer (i.e., E$^2$MetaNODE). Next, we introduce them in details.

\subsection{Computation Complexity Analysis of MetaNODE}
\label{sec_5_1}
\noindent {\bf Theoretical Complexity Analysis.} As shown in Section~\ref{section_3_3}, the proposed GradNet in MetaNODE consists of $H$ inference modules and each module contains a gradient estimator and a weight aggregator. Next, we attempt to analyze its computation complexity from theory: (1) as shown in Eq.~\ref{eq12}, the gradient estimator mainly calculate the difference vector between each support/query sample $x \in \mathcal{S}\cup\mathcal{Q}$ and each prototype $k \in [0, N-1]$. Thus, its computation complexity around is $O\left[N\left(K+M\right)N\right]$ where $N(K+M)$ is the number of support/query samples; (2) as shown in Eqs.~\ref{eq13} and \ref{eq17}, the computation complexity mainly lies in the relation layer $g_{\theta_{gr}^{l}}()$ with multi-head based attention mechanism. Thus, its computation complexity is around $O\left\{\left[N\left(K+M\right)\right]^2dN\right\}$ where $d$ denotes the embedding dimension of the embedding layer $g_{\theta_{ge}^{l}}()$; (3) as shown in Eq.~\ref{eq19}, we finally ensemble the estimation from all inference modules, thus the overall computation complexity of our GradNet is around $O\left\{\big\{N\left(K+M\right)N+\left[N\left(K+M\right)\right]^2dN\big\}H\right\}$.

From this above theoretical complexity, we can see that the computation cost of our GradNet in MetaNODE is very big, which grows in a binomial trend as the number $N$ of classes and the number $K$ of samples per class increase.

\noindent {\bf Experimental Cost Analysis.} To further analyze the computation efficiency of our MetaNODE, we select some related FSL methods (i.e., ClassifierBaseline \cite{Yinbo20}, ALFA \cite{BaikCCKL20}, and CloserLook \cite{ChenLKWH19}) as our baselines and report the test time of these baseline methods and our MetaNODE. All results are averaged on 600 episodes and counted on the same machine with an NVIDIA RTX8000 GPU, which are reported in Table~\ref{table222}. From these experimental results, we can find that 1) compared with ClassifierBaseline, our MetaNODE method (employing RK4 as our ODE solver) takes relatively longer time, around 10 times. This imply that the process of prototype optimization is very inefficient in our MetaNODE method; 2) the running time of our MetaNODE (employing RK4 as our ODE solver) exceed existing optimization-based FSL methods (i.e., ALFA \cite{BaikCCKL20} and CloserLook \cite{ChenLKWH19}) by a large margin. This further show the inefficiency drawback of our MetaNODE; 3) the running time of our MetaNODE drops significantly, around 2/3, when we replace the RK4 ODE solver by Euler ODE solver, which means that employing a low-order ODE solver are helpful for improving computation efficiency but it will degread the FSL performance (see Table~\ref{table88}); and 4) the running time of our MetaNODE drops significantly by a large margin, when we simplify our GradNet by removing attention and multi-module ensemble mechanisms. This also verifies our theoretical analysis.

Based on these above theoretical and experiment analysis, we find that the GradNet and ODE solver are key factors impacting computing efficiency of our MetaNODE. Next, we attempt to improve them by simplifying our GradNet and designing a faster and efficient ODE solver, respectively. 

\subsection{Theoretic Analysis of GradNet}

To understand how our GradNet designed in Section~\ref{section_3_3} works, in this subsection, we provide a brief theoretic analysis on the GradNet. Intuitively, the GradNet aims to estimate the gradient flow of prototype optimization at each time step. Next, we take the nearest-prototype cosine classifier shown in Eq.~\ref{eq7} to analyze our GradNet theoretically.

\noindent {\bf Proposition 1.} the Euclidean distance is equivalent to cosine similarity when two sample points $A$ and $B$ are normlized into the same spherical space with radius $r$.

\noindent \emph{Proof.}
We take two image features in 2D space as an example to prove the Proposition 1. Let us denote these two image features by using $A=(z_1^1, z_1^2)$ and $B=(z_2^1, z_2^2)$. We normalize these two points into a spherical space with radius $r$. As a result, these two normalized sample points can be obtained, i.e., $A'=(\frac{z_1^1}{\sqrt{{z_1^2}^2 + {z_1^2}^2}}, \frac{z_1^2}{\sqrt{{z_1^1}^2 + {z_1^2}^2}})$ and $B'=(\frac{z_2^1}{\sqrt{{z_2^1}^2 + {z_2^2}^2}}, \frac{z_2^2}{\sqrt{{z_2^1}^2 + {z_2^2}^2}})$. 

Then, we can calculate the cosine similarity between these two normalized points $A'$ and $B'$. That is, 
\begin{small}
\begin{equation}
	\begin{aligned}
		\mathop{cos}(A', B') &= \frac{\frac{z_1^1}{\sqrt{{z_1^1}^2 + {z_1^2}^2}}  \frac{z_2^1}{\sqrt{{z_2^1}^2 + {z_2^2}^2}} + \frac{z_1^2}{\sqrt{{z_1^1}^2 + {z_1^2}^2}} \frac{z_2^2}{\sqrt{{z_2^1}^2 + {z_2^2}^2}}}{\|A'\|\|B'\|},
	\end{aligned}
	\label{eq13_}
\end{equation}
\end{small}where the two points $A'$ and $B'$ are normalized into a spherical space with radius $r$, thus the Eq.~\ref{eq13_} can be simplified as:
\begin{small}
	\begin{equation}
		\begin{aligned}
			\mathop{cos}(A', B') &= \frac{\frac{z_1^1}{\sqrt{{z_1^1}^2 + {z_1^2}^2}}  \frac{z_2^1}{\sqrt{{z_2^1}^2 + {z_2^2}^2}} + \frac{z_1^2}{\sqrt{{z_1^1}^2 + {z_1^2}^2}} \frac{z_2^2}{\sqrt{{z_2^1}^2 + {z_2^2}^2}}}{r^2}.
		\end{aligned}
		\label{eq14_}
	\end{equation}
\end{small}Besides, we can also calculate the Euclidean distance between these two normalized points $A'$ and $B'$. That is,
\begin{scriptsize}
    \begin{equation}
		\begin{aligned}
			&\mathop{euc}(A', B')\\ &= \sqrt{(\frac{z_1^1}{\sqrt{{z_1^1}^2 + {z_1^2}^2}}-\frac{z_2^1}{\sqrt{{z_2^1}^2 + {z_2^2}^2}})^2+(\frac{z_1^2}{\sqrt{{z_1^1}^2 + {z_1^2}^2}}-\frac{z_2^2}{\sqrt{{z_2^1}^2 + {z_2^2}^2}})^2},
		\end{aligned}
		\label{eq15_}
    \end{equation}
\end{scriptsize}where the two points $A'$ and $B'$ are normalized into a spherical space with radius $r$, thus the Eq.~\ref{eq15_} can be simplified as:
\begin{scriptsize}
    \begin{equation}
		\begin{aligned}
			&\mathop{euc}(A', B')\\ &=\sqrt{2r^2-2(\frac{z_1^1}{\sqrt{{z_1^1}^2 + {z_1^2}^2}}  \frac{z_2^1}{\sqrt{{z_2^1}^2 + {z_2^2}^2}} + \frac{z_1^2}{\sqrt{{z_1^1}^2 + {z_1^2}^2}} \frac{z_2^2}{\sqrt{{z_2^1}^2 + {z_2^2}^2}})} \\ &=\sqrt{2r^2-2r^2\mathop{cos}(A', B')}.
		\end{aligned}
		\label{eq16_}
    \end{equation}
\end{scriptsize}That is, Euclidean distance and cosine similarity between $A'$ and $B'$ satisfy $\mathop{euc}(A', B')=\sqrt{2r^2-2r^2\mathop{cos}(A', B')}$. This means that the relation of Euclidean distance with normalization and cosine similarity between two normalized points $A'$ and $B'$ is monotonical. In other words, the Euclidean distance with normalization is positively related to cosine similarity when two sample points $A$ and $B$ are normalized into a spherical space with radius $r$.

Based on {\bf Proposition 1}, we can resort to the nearest-prototype classifier with Euclidean distance to analyze the prototype optimization dynamic of the nearest-prototype cosine classifier from the perspective of gradient. Formally, given a labeled sample $i$, its features and its one-hot label are denoted by $f_{\theta_f}(x_i)$ and $y_i={y_i^1, ..., y_i^k, ..., y_i^N}$ ($y_i^k=0/1$), we can calculate its cross-entropy loss $L(y_i, \hat{y}_i)$ as follows:
\begin{equation}
	\begin{aligned}
		\mathop{L}(y_i, \hat{y}_i) &= - \sum_{k=1}^{N} y_i^{k} log\ P(y=k|x_i),
	\end{aligned} 
	\label{eq17_}
\end{equation}where $P(y=k|x_i)$ denotes the probability of the sample $x_i$ belonging to class $k$. Following Eq.~
\ref{eq7}, we compute the probability $P(y=k|x_i)$ by using a cosine classifier. That is,
\begin{equation}
	P(y=k|x_i) = \frac{e^{\gamma \cdot \mathop{cos}(f_{\theta_f}(x_i), p_{k})}}{\sum_c e^{\gamma \cdot \mathop{cos}(f_{\theta_f}(x_i), p_{c})}}.
	\label{eq18_}
\end{equation}Then, based on {\bf Proposition 1}, we can transform Eq.~\ref{eq18_} into the former with Euclidean distance by normalizing the sample feature $f_{\theta_f}(x_i)$ into the spherical space with radius $\|p_{k}\|$. That is,
\begin{equation}
	\begin{aligned}
	P(y=k|x_i) &= \frac{e^{-\gamma \cdot \mathop{euc}(f_{\theta_f}(x_i), p_{k})}}{\sum_c e^{\gamma \cdot \mathop{euc}(f_{\theta_f}(x_i), p_{c})}} \\ &= \frac{e^{-\gamma \cdot (f_{\theta_f}(x_i) - p_{k})^2}}{\sum_c e^{\gamma \cdot (f_{\theta_f}(x_i) - p_{c})^2}}.
    \end{aligned}
	\label{eq19_}
\end{equation}Thus, we can summary the cross entropy loss $\mathop{L}(y_i, \hat{y}_i)$ as:
\begin{equation}
	\begin{aligned}
		\mathop{L}(y_i, \hat{y}_i) &= - \sum_{k=1}^{N} y_i^{k} log\ P(y=k|x_i), \\ P(y=k&|x_i) = \frac{e^{-\gamma \cdot (f_{\theta_f}(x_i) - p_{k})^2}}{\sum_c e^{\gamma \cdot (f_{\theta_f}(x_i) - p_{c})^2}}.
	\end{aligned} 
	\label{eq20_}
\end{equation}

By the chain rule of gradients, we can compute the gradient $\frac{\partial L(y_i, \hat{y}_i)}{\partial p_k}$ of prototype $\{p_k\}_{k=1}^{N}$ as follows:
\begin{equation}
	\begin{aligned}
		\frac{\partial L(y_i, \hat{y}_i)}{\partial p_k} = (P(y=k|x_i)-y_i^k)(f_{\theta_f}(x_i) - p_{k}).
	\end{aligned} 
	\label{eq21_}
\end{equation}Note that Eq.~\ref{eq21_} only shows the case of a labeled sample. When $M$ samples are provided, the gradient $\frac{\partial L}{\partial p_k}$ of prototype $\{p_k\}_{k=1}^{N}$ can be expressed as:
\begin{equation}
	\begin{aligned}
		\frac{\partial L}{\partial p_k} = \frac{1}{M} \sum_{i=1}^{M}(P(y=k|x_i)-y_i^k)(f_{\theta_f}(x_i) - p_{k}).
	\end{aligned} 
	\label{eq22_}
\end{equation}
Finally, based on the Eq.~\ref{eq_7777}, the gradiend flow of prototype at time $t$ can be summaried as follows:
\begin{equation}
	\begin{aligned}
			\frac{\mathrm{d} p_k(t)}{\mathrm{d} t} &= - \nabla L(p_k(t)) \\&= - \frac{\partial L}{\partial p_k(t)} \\ &= \frac{1}{M} \sum_{i=1}^{M}\underbrace{(y_i^k-P(y=k|x_i, t))}_{Part\ 1}\overbrace{(f_{\theta_f}(x_i) - p_{k}(t))}^{Part\ 2}.
	\end{aligned} 
	\label{eq23_}
\end{equation}

Let us revisit the design details of gradient flow inference network, i.e., Eqs.\ref{eq12} $\sim$ \ref{eq17}. Among them, Eq.\ref{eq12} accounts for estimating gridient flow, Eq.\ref{eq13} accounts for estimating weights, and Eq.\ref{eq17} aims to aggregate all gridients for all sample in a weighted sum manner. Comparing with Eqs.\ref{eq12} $\sim$ \ref{eq17} and Eq.\ref{eq23_}, we can find that the Eq.\ref{eq12} actually correspondings to the part 2 of Eq.\ref{eq23_}. Here, the scale layer $g_{\theta_{gs}^{l}}()$ with parameters $\theta_{gs}^{l}$ aims to normalize the sample $x_i$ into an approproate spherical space for transforming cosine similarity into Euclidean distance. The Eq.\ref{eq13} actually correspondings to the part 1 of Eq.\ref{eq23_}, which aims to estimate the contributation weights for each sample. The Eq.\ref{eq17} actually correspondings to the overall Eq.\ref{eq23_}.

In conclusion, the estimation of the part 1 in Eq.\ref{eq23_} is a key challenge since the label $y_i$ is unknown for each query samples and inaccuracy for each support sample. In our conference version, we design a multi-head attention mechanism to estimate of the part 1 in Eq.\ref{eq23_}, which consumes a huge amount of computation cost as analyzed in Section~\ref{sec_5_1}. Next, we introduce an effective and efficient version to estimate the part 1 in Eq.\ref{eq23_} by using a ResNet-like network, as shown in Figure~\ref{fig311}.

	

\subsection{Efficient and Effective GradNet (E$^2$GradNet)}
Based on the above theoretic analysis of GradNet, from Eq.~\ref{eq23_}, we can see that the key of estimating gradient flow $\frac{\mathrm{d} p_k(t)}{\mathrm{d} t}$ of prototype is how to calculate the part 1 (i.e., $(y_i^k-P(y=k|x_i, t)$). Here, $P(y=k|x_i, t)$ denotes the estimation of class probability by using the prototype estimation $p_k(t)$ at time $t$. As shown in Figure~\ref{fig311}, following Eq.~\ref{eq7}, we use cosine classifier to estimate the class probability:
\begin{equation}
	P(y=k|x_i, t) = \frac{e^{\gamma \cdot <f_{\theta_f}(x_i), p_{k}(t)>}}{\sum_c e^{\gamma \cdot <f_{\theta_f}(x_i), p_{c}(t)>}},
\end{equation}
where $<\cdot>$ denotes the cosine similarity, and $\gamma$ is a scale parameter. Besides, in the part 1 (i.e., $(y_i^k-P(y=k|x_i, t)$), $y_i^k$ denotes the ground true class probability of sample $x_i$ belonging to class $k$ (i.e., $y_i^k = 1$ if the sample $x_i$ belongs to class $k$ otherwise $y_i^k = 0$). Since this term $y_i^k$ is unknown for each query sample $x \in \mathcal{Q}'$, but it is correlative to the estimation $P(y=k|x_i, t)$ of class probability, thus we take the estimation $P(y=k|x_i, t)$ of class probability as inputs and then employ a simple neural network $f_{\theta_ge}()$ with parameters $\theta_ge$ to estimate $y_i^k$. That is,
\begin{equation}
	y_i^k = f_{\theta_ge}(P(y=k|x_i, t)), k = 0,1,...,N-1.
\end{equation}
The estimation of part 1 (i.e., $(y_i^k-P(y=k|x_i, t)$) looks like a ResNet, which can be expressed as follows:
\begin{equation}
	\begin{aligned}
	y_i^k-P(y=k|x_i, t) = &f_{\theta_ge}(P(y=k|x_i, t)) - P(y=k|x_i, t), \\ &k = 0,1,...,N-1.
	\end{aligned}
 \label{eq26}
\end{equation}In our paper, we implement the $f_{\theta_{ge}}()$ by using a two-layer multilayer perceptron (MLP). 

\noindent {\bf Theoretical Complexity.} As shown in Eq.~\ref{eq26}, different from the proposed GradNet in MetaNODE, our $E^2$GradNet does not require calculating computation-complex multi-head attention, such that its computation complexity only stems from the difference vector calculation between each support/query sample $x \in \mathcal{S}\cup\mathcal{Q}$ and each prototype $k \in [0, N-1]$. Thus, the overall complexity analysis only is around $O\left[N\left(K+M\right)N\right]$. Compared with the proposed GradNet in MetaNODE $O\left\{\big\{N\left(K+M\right)N+\left[N\left(K+M\right)\right]^2dN\big\}H\right\}$, our $E^2$GradNet improves computational efficiency significantly (please refer to Table~\ref{table221} for more details).

\begin{table*}
	\caption{Experiment results on miniImageNet and tieredImageNet. The best results are highlighted in bold. ``CV'' denotes conference version.}
	\begin{center}
		\smallskip\scalebox
		{0.95}{
			\begin{tabular}{l|l|c|c|c|c|c|c}
				\hline
				\multicolumn{1}{l|}{\multirow{2}{*}{Setting}}&\multicolumn{1}{c|}{\multirow{2}{*}{Method}}&\multicolumn{1}{c|}{\multirow{2}{*}{Type}}&\multicolumn{1}{c|}{\multirow{2}{*}{Backbone}}& \multicolumn{2}{c|}{miniImagenet} & \multicolumn{2}{c}{tieredImagenet} \\ 
				\cline{5-8}
				& & & & 5-way 1-shot & 5-way 5-shot & 5-way 1-shot & 5-way 5-shot \\
				\hline \hline
				\multicolumn{1}{l|}{\multirow{18}{*}{\makecell*[c]{Transductive}}} 
				& TPN\cite{LiuLPKYHY19} & Graph & Conv4 &  $55.51 \pm 0.86\%$  & $69.86 \pm 0.65\%$ & $59.91 \pm 0.94\%$  & $73.30 \pm 0.75\%$ \\
				& TEAM \cite{Qiao000HW19} & Graph & ResNet18 &  $60.07 \%$  & $75.90 \%$ & $-$  & $-$ \\
				& EdgeLabel\cite{kim2019edge} & Graph & ResNet12 &  $-$  & $ 76.37 \%$ & $-$  & $80.15 \%$ \\
				& DPGN \cite{YangLZZZL20} & Graph & ResNet12 &  $67.77 \pm 0.32\%$  & $84.60 \pm 0.43\%$ & $72.45 \pm 0.51\%$  & $87.24 \pm 0.39\%$ \\
				& EPNet \cite{RodriguezLDL20} & Graph & ResNet12 & $66.50 \pm 0.89\%$  & $81.06 \pm 0.60\%$ & $76.53 \pm 0.87\%$  & $87.32 \pm 0.64\%$ \\
				& MCGN \cite{Tang_2021_CVPR} & Graph & Conv4 & $67.32 \pm 0.43\%$  & $83.03 \pm 0.54\%$ & $71.21 \pm 0.85\%$  & $85.98 \pm 0.98\%$ \\
				& ECKPN \cite{Chen_2021_CVPR} & Graph & ResNet12 & $70.48 \pm 0.38\%$  & 85.42 $\pm$ 0.46 $\%$ & $73.59 \pm 0.45\%$  & 88.13 $\pm$ 0.28$\%$ \\
				& TFT \cite{DhillonCRS20} & Pre-training & WRN-28-10 &  $65.73 \pm 0.68\%$  & $78.40 \pm 0.52\%$ & $73.34 \pm 0.71\%$  & $85.50 \pm 0.50\%$ \\
				& ICI \cite{WangXLZF20} & Pre-training& ResNet12 & $65.77 \%$  & $78.94 \%$ & $80.56 \%$  & $87.93 \%$ \\
				& TIM-GD \cite{Malik20} & Pre-training & ResNet18 & $73.9 \%$  & 85.0\% & $79.9 \%$  & 88.5 \% \\
				& LaplacianShot \cite{ZikoDGA20} & Pre-training & ResNet18 & $72.11 \pm 0.19\%$  & $82.31 \pm 0.14\%$ & $78.98 \pm 0.21\%$  & $86.39 \pm 0.16\%$ \\
				& SIB \cite{HuMXSOLD20} & Pre-training & WRN-28-10 & $70.0 \pm 0.6\%$  & $79.2 \pm 0.4\%$ & -  & - \\
                    & protoLP \cite{zhu2023transductive} & Pre-training & ResNet-18 & $75.77 \pm 0.29\%$  & $84.00 \pm 0.16\%$ & $82.32 \pm 0.27\%$   & $88.09 \pm 0.15\%$ \\  
				& BD-CSPN \cite{YaohuiWang_pr} & Pre-training & WRN-28-10 &  $70.31 \pm 0.93\%$  & $81.89 \pm 0.60\%$ & $78.74 \pm 0.95\%$  & $86.92 \pm 0.63\%$ \\	
				& SRestoreNet \cite{XueW20} & Pre-training & ResNet18 &  $61.14 \pm 0.22\%$  & - & -  & - \\	
				& ProtoComNet \cite{zhang2021} & Pre-training & ResNet12 &  $73.13 \pm 0.85\%$  & $82.06 \pm 0.54\%$ & $81.04 \pm 0.89\%$  & $87.42 \pm 0.57\%$ \\
				\cline{2-8}
				& MetaNODE (CV) & Pre-training & ResNet12 & 77.92$\pm$ 0.99$\%$ & 85.13 $\pm$ 0.62$\%$ & 83.46 $\pm$ 0.92$\%$ & 88.46 $\pm$ 0.57$\%$ \\	
				& E$^2$MetaNODE (ours) & Pre-training & ResNet12 & \emph{\textbf{78.62}} $\pm$ \emph{\textbf{1.00}}$\%$ & \emph{\textbf{85.45}} $\pm$ \emph{\textbf{0.56}}$\%$ & \emph{\textbf{84.10}} $\pm$ \emph{\textbf{0.96}}$\%$ & \emph{\textbf{88.93}} $\pm$ \emph{\textbf{0.60}}$\%$ \\	
				\hline
				\multicolumn{1}{l|}{\multirow{20}{*}{\makecell*[c]{Inductive}}} 
				& CTM \cite{li2019finding} & Metric & ResNet18 &  $62.05 \pm 0.55\%$  & $78.63 \pm 0.06\%$ & $64.78 \pm 0.11\%$  & $81.05 \pm 0.52\%$ \\
				& CAN \cite{HouCMSC19} & Metric & ResNet12 &  $63.85 \pm 0.48\%$  & $79.44 \pm 0.34\%$ & $69.89 \pm 0.51\%$  & $84.23 \pm 0.37\%$ \\
                & MetaCA \cite{gao2022curvature} & Metric & ResNet12 & 63.13 $\pm$ 0.41$\%$ & 81.04 $\pm$ 0.39$\%$ & 68.49 $\pm$ 0.56$\%$ & 83.84 $\pm$ 0.40$\%$ \\
				& MAML \cite{FinnAL17} & Optimization & ResNet12 & $58.37 \pm 0.49\%$  & $69.76 \pm 0.46\%$ & $58.58 \pm 0.49\%$  & $71.24 \pm 0.43\%$ \\
				& iMAML\cite{RajeswaranFKL19} & Optimization & Conv4 & $49.30 \pm 1.88\%$  & $-$ & $-$  & $-$ \\
				& Warp-MAML \cite{FlennerhagRPVYH20} & Optimization & Conv4 & $52.3 \pm 0.8\%$  & $68.4 \pm 0.6\%$ & $57.2 \pm 0.9\%$  & $71.4 \pm 0.7\%$ \\
				& ALFA \cite{BaikCCKL20} & Optimization & ResNet12 & $59.74 \pm 0.49\%$  & $77.96 \pm 0.41\%$ & $64.62 \pm 0.49\%$  & $82.48 \pm 0.38\%$ \\
				& sparse-MAML \cite{Johannes21} & Optimization & Conv4  & $56.39 \pm 0.38\%$  & $73.01 \pm 0.24\%$ & $-$  & $-$ \\
                 & GAP \cite{kang2023meta} & Optimization & Conv4 & $54.86 \pm 0.85\%$  & $71.55 \pm 0.61\%$ & $57.60 \pm 0.93\%$  & $74.90 \pm 0.68\%$ \\
                & MeTAL \cite{baik2021meta} & Optimization & ResNet12 & 59.64 $\pm$ 0.38$\%$ & 76.20 $\pm$ 0.19$\%$ &  63.89 $\pm$ 0.43$\%$ & 80.14 $\pm$ 0.40$\%$ \\
                
				& Neg-Cosine \cite{LiuCLL0LH20} & Pre-training & ResNet12 &  $63.85 \pm 0.81\%$  & $81.57 \pm 0.56\%$ & -  & - \\
				& P-Transfer \cite{ShenLQSC21} & Pre-training & ResNet12 &  $64.21\pm 0.77\%$  & $80.38 \pm 0.59\%$ & -  & - \\
				& Meta-UAFS \cite{Zhiuncertainty21} & Pre-training & ResNet12 &  $64.22\pm 0.67\%$  & $79.99 \pm 0.49\%$ & $69.13 \pm 0.84\%$  & $84.33 \pm 0.59\%$ \\
				& RestoreNet \cite{XueW20} & Pre-training & ResNet12 &  $59.28 \pm 0.20\%$  & - & -  & - \\
				& ClassifierBaseline \cite{Yinbo20} & Pre-training & ResNet12 & 61.22 $\pm$ 0.84$\%$ & 78.72 $\pm$ 0.60$\%$ & 69.71 $\pm$ 0.88$\%$ & 83.87 $\pm$ 0.64$\%$ \\
                & MetaQDA \cite{zhang2021shallow} & Pre-training & ResNet12 & 65.12 $\pm$ 0.66$\%$ & 80.98 $\pm$ 0.75$\%$ & 69.97 $\pm$ 0.52$\%$ & 85.51 $\pm$ 0.58$\%$ \\
                & ProtoDiff \cite{du2023protodiff} & Pre-training & ResNet12 & 66.63 $\pm$ 0.21$\%$ & 83.48 $\pm$ 0.15$\%$ & 72.95 $\pm$ 0.24$\%$ & 85.15 $\pm$ 0.18$\%$ \\
				%
    
				\cline{2-8}	
				& MetaNODE (CV) & Pre-training & ResNet12 & 66.07 $\pm$ 0.79$\%$ & 81.93 $\pm$ 0.55$\%$ & 72.72 $\pm$ 0.90$\%$ & 86.45 $\pm$ 0.62$\%$ \\	
				& E$^2$MetaNODE (ours) & Pre-training & ResNet12 & 66.37 $\pm$ 0.80$\%$ & 81.97 $\pm$ 0.56$\%$ & 73.16 $\pm$ 0.92$\%$ & 86.66 $\pm$ 0.62$\%$ \\			
				\hline
		\end{tabular}}
	\end{center}
        \vspace{-15pt}
	\label{table1}
\end{table*} 

\subsection{Efficient and Effective ODE Solver (E$^2$Solver)} 
In previous conference version, we employ various ODE solvers, including Euler and RK4 solvers, to solve the optimal prototypes for FSL. Theoretically, the RK4 solver has higher integration accuracy  meanwhile also consumes more compution cost than the Euler solver (see Table~\ref{table221}). The main reason is because RK4 solver uses a lot of computing resources to ensure more high integral order, such that the lower truncation error can be obtained for sloving ODE. Next, we derived the truncation error for Euler and RK4 solvers, respectively. For the Euler, the estimate $y_{n+1}$ of state at $n+1$ can be expressed as follows:
\begin{equation}
	\begin{aligned}
		y_{n+1} = y(x_n) + h f(x_n, y(x_n)).
	\end{aligned}
\label{eq_27_}
\end{equation}where $h$ is integral step and $f(x_n, y(x_n))$ is gradient flow of state $y(x_n)$ at $x_n$. For it ground state $y(x_{n+1})$, we can obtain it by applying taylor expansion at $x_n$. That is, 
\begin{equation}
	\begin{aligned}
		y(x_{n+1}) = y(x_n) + h y'(x_n) + \frac{h^2}{2}y''(x_n)+\mathcal{O}(h^3).
	\end{aligned}
\end{equation}Then, we can compute its truncation error as follows:
\begin{equation}
	\begin{aligned}
		y(x_{n+1})-y_{n+1} = \frac{h^2}{2}y''(x_n)+\mathcal{O}(h^3) \approx \mathcal{O}(h^2).
	\end{aligned}
\label{eq_29_}
\end{equation}  For the RK4, the estimate $y_{n+1}$ of state at $n+1$ can be expressed as follows:
\begin{equation}
\begin{aligned}
	y_{n+1} = y(x_n) + &\frac{h}{6} (K_1 + 2 K_2 + 2 K_3 + K_4) \\ K_1 &= f(x_n, y(x_n)), \\K_2 = &f(x_{n+\frac{1}{2}}, y_n + \frac{h}{2} K_1), \\K_3 = &f(x_{n+\frac{1}{2}}, y_n + \frac{h}{2} K_2), \\K_4 = &f(x_{n+1}, y_n + h K_3),.
\end{aligned}
\label{eq_30_}
\end{equation} Simliarly, we can also compute the truncation error of the RK4 solver by using taylor expansion, that is,
\begin{equation}
	\begin{aligned}
		y(x_{n+1})-y_{n+1} = \frac{h^4}{24}y''''(x_n)+\mathcal{O}(h^5) \approx \mathcal{O}(h^4).
	\end{aligned}
\label{eq_31_}
\end{equation}Finally, from Eq.~\ref{eq_27_} $\sim$ Eq.~\ref{eq_31_}, we can see that the RK4 solver apples more the forward calculation of gradient flow for reducing the truncation error. In our MetaNODE optimizer, the estimation of gradient flow is implemented by using GradNet and performing multiple forward propagation of GradNet will significantly reduce the efficiency. 


To address this issue, instead of using multiple forward propagation of GradNet to reduce the truncation error (i.e. adopting RK4 solver), we propose to employ a simple neural network $\eta_{\theta_{\eta}}()$ with parameters $\theta_{\eta}$ to directly estimate the truncation error  $\mathcal{O}(h^2)$ for the Euler solver. That is,
\begin{equation}
	\begin{aligned}
		y_{n+1} = y(x_n) + h f(x_n, y(x_n)) + \eta_{\theta_{\eta}}(x_n, y(x_n)).
	\end{aligned}
	\label{eq_32_}
\end{equation}The advantage of such design is that our ODE solver not only owns high efficiency like Euler but also a lower truncation error than Euler for solving Neural ODE. Note that in extension version, our E$^2$MetaNODE consists of E$^2$GradNet and E$^2$Solver, which are together trained in an episodic training strategy describe in meta-training phase of Figure~\ref{fig2}.


\begin{table}
	\caption{Experiment results on CUB-200-2011.}
	\begin{center}
		\smallskip\scalebox
		{0.9}{
			\begin{tabular}{c|l|c|c}
				\hline
				\multicolumn{1}{l|}{\multirow{2}{*}{Setting}}&\multicolumn{1}{c|}{\multirow{2}{*}{Method}}& \multicolumn{2}{c}{CUB-200-2011} \\ 
				\cline{3-4}
				& & 5-way 1-shot & 5-way 5-shot \\
				\hline \hline
				\multicolumn{1}{l|}{\multirow{7}{*}{\makecell*[c]{Trans\\ductive}}}
				& TEAM \cite{Qiao000HW19}&  $80.16\%$  & $87.17\%$ \\
				& DPGN \cite{YangLZZZL20}& $75.71 \pm 0.47\%$  & $91.48 \pm 0.33\%$\\
				& EPNet \cite{RodriguezLDL20}&$82.85 \pm 0.81\%$  &$91.32 \pm 0.41\%$\\
				& ECKPN \cite{Chen_2021_CVPR}&$77.43 \pm 0.54\%$  &$92.21 \pm 0.41\%$\\
				& ICI \cite{WangXLZF20}& $87.87\%$  & $92.38\%$ \\
				& LaplacianShot \cite{ZikoDGA20} & $80.96\%$  & $88.68\%$ \\
				& BD-CSPN \cite{YaohuiWang_pr}&  $87.45\%$  & $91.74\%$ \\
				& RestoreNet \cite{XueW20}&  $76.85 \pm 0.95\%$  & - \\		
				\cline{2-4}
				& MetaNODE (CV) & 90.94 $\pm$ 0.62$\%$ & 93.18 $\pm$ 0.38$\%$ \\	
				& E$^2$MetaNODE (ours) & \emph{\textbf{91.27}} $\pm$ \emph{\textbf{0.67}}$\%$ & \emph{\textbf{93.80}} $\pm$ \emph{\textbf{0.35}}$\%$ \\		
				\hline	
				\multicolumn{1}{l|}{\multirow{5}{*}{\makecell*[c]{In\\ductive}}} &
				
				RestoreNet \cite{XueW20}&  $74.32 \pm 0.91\%$  & - \\
				& Neg-Cosine \cite{LiuCLL0LH20}&  $72.66 \pm 0.85\%$  & $89.40 \pm 0.43\%$ \\
				& P-Transfer  \cite{ShenLQSC21}&  $73.88 \pm 0.87\%$  & $87.81 \pm 0.48\%$ \\
				& ClassifierBaseline \cite{Yinbo20}&  $74.96 \pm 0.86\%$  & $88.89 \pm 0.43\%$ \\
				\cline{2-4}
				& MetaNODE (CV) & 80.82 $\pm$ 0.75$\%$ & 91.77 $\pm$ 0.49$\%$ \\	
                    & E$^2$MetaNODE (ours) & \textbf{83.36} $\pm$ \textbf{0.71}$\%$ & \textbf{92.96} $\pm$ \textbf{0.36}$\%$ \\	
				\hline
		\end{tabular}}
	\end{center}
	\label{table2}
\end{table}

\section{Performance Evaluation}
\label{section_5}
\subsection{Datasets and Settings}
\noindent \textbf{MiniImagenet.} The dataset consists of 100 classes, where each class contains 600 images. Following the standard split in \cite{Yinbo20}, we split the data set into 64, 16, and 20 classes for training, validation, and test, respectively. 

\noindent \textbf{TieredImagenet.}
The dataset is a larger dataset with 608 classes. Each class contains 1200 images. Following \cite{Yinbo20}, the dataset is split into 20, 6, and 8 high-level semantic classes for training, validation, and test, respectively.  

\noindent \textbf{CUB-200-2011.}
The dataset is a fine-grained bird recognition dataset with 200 classes. It contains about 11,788 images. Following the standard split in \cite{ChenLKWH19}, we split the data set into 100 classes, 50 classes, and 50 classes for training, validation, and test, respectively. 

\subsection{Implementation Details}
\noindent \textbf{Network Details.}
We use ResNet12 \cite{Yinbo20} as the feature extractor. 
For our MetaNODE conference version, in GradNet, we use four inference modules to estimate the gradient flow. For each module, we use a two-layer MLP with 512-dimensional hidden layer for the scale layer, a single-layer perceptron with 512-dimensional outputs for the the embedding layer, a multi-head attention module with $8$ heads and each head contains 16 units for the relation layer, and a single-layers perceptron and a softmax layer for the output layer. ELU \cite{ClevertUH15} is used as the activation function.

\begin{table}	
	\centering
	\caption{Experiments of prototype bias on on miniImagenet, TieredImagenet, and CUB-200-2011 datasets. }
	\smallskip\scalebox
	{1.0}{
		\smallskip\begin{tabular}{l|c|c}
			\hline
			Methods & Initial Prototypes & Optimal Prototypes \\
			\hline \hline
			\multicolumn{3}{c}{\multirow{1}{*}{MiniImagenet}} \\
			\hline
			BD-CSPN & 0.64 & 0.75 \\
			SRestoreNet & 0.64 & 0.85 \\
			ProtoComNet & 0.64 & 0.91 \\
			\hline
			MetaNODE & 0.64 & 0.93\\
			E$^2$MetaNODE & 0.64 & 0.94 \\
			\hline \hline
			\multicolumn{3}{c}{\multirow{1}{*}{TieredImagenet}} \\
			\hline
			BD-CSPN & 0.71 & 0.83 \\
			SRestoreNet & 0.71 & 0.91 \\
			ProtoComNet & 0.71 & 0.95 \\
			\hline
			MetaNODE & 0.71 & 0.96 \\
			E$^2$MetaNODE & 0.71 & 0.97 \\
			\hline \hline
			\multicolumn{3}{c}{\multirow{1}{*}{CUB-200-2011}} \\
			\hline
			BD-CSPN & 0.69 & 0.79 \\
			SRestoreNet & 0.69 & 0.89 \\
			ProtoComNet & 0.69 & 0.95 \\
			\hline
			MetaNODE & 0.69 & 0.96 \\
			E$^2$MetaNODE & 0.69 & 0.97 \\
			\hline \hline
	\end{tabular}}\smallskip
	\label{table3}
\end{table}

For our E$^2$MetaNODE extension version, we employ a two-layer MLP with $N$-dimensional input layer, $N$-dimensional hidden layer, and $N$-dimensional output layer for the ResNet-like gradient flow inference network $\eta_{\theta_{\eta}}()$. For meta-learning-based ODE solver, we use a two-layer MLP with 512-dimensional input layer, 512-dimensional hidden layer, and 512-dimensional output layer to implement and Euler ODE solve is used as base ODE solver.

\noindent \textbf{Training details.}
Following \cite{Yinbo20}, we use an SGD optimizer to train the feature extractor for 100 epochs. 
In the meta-training phase, we train our MetaNODE and E$^2$MetaNODE meta-optimizer 50 epochs using Adam with a learning rate of 0.0001 and a weight decay of 0.0005, where the learning rate is decayed by 0.1 at epochs 15, 30, and 40, respectively. 

\noindent \textbf{Evaluation.} Following \cite{Yinbo20}, we evaluate our method on 600 randomly sampled episodes (5-way 1/5-shot tasks) from the novel classes and report the mean accuracy together with the 95\% confidence interval. In each episode, we randomly sample 15 images per class as the query set.



\subsection{Performance Evaluation}
We evaluate our MetaNODE and E$^2$MetaNODE prototype optimization framework and various state-of-the-art methods on general and fine-grained few-shot image recognition tasks. Among these methods, ClassifierBaseline, RestoreNet, BD-CSPN, SIB, ProtoComNet, SRestoreNet, and ProtoDiff are our strong competitors since they also focus on learning reliable prototypes for FSL. However, different from these one-step or discrete-time rectification methods, our MetaNODE and E$^2$MetaNODE focuses on rectifing prototype in a continuous-time optimization manner. In particular, ProtoComNet introduces external knowledge for FSL.

\begin{table}
	\centering
	\caption{Experiments of gradient bias on miniImagenet, TieredImagenet, and CUB-200-2011 datasets. }
	\smallskip\scalebox
	{1.0}{
		\smallskip\begin{tabular}{l|c|c}
			\hline
			Methods & Averaged Gradient & Infered Gradient \\
			\hline \hline
			\multicolumn{3}{c}{\multirow{1}{*}{MiniImagenet}} \\
			\hline
			SIB & 0.0441 & 0.0761 \\
			MetaNODE & 0.0441 & 0.1701 \\
			E$^2$MetaNODE & 0.0441 & 0.1854 \\
			\hline \hline
			\multicolumn{3}{c}{\multirow{1}{*}{TieredImagenet}} \\
			\hline
			SIB & 0.1230 & 0.2225 \\
			MetaNODE & 0.1230 & 0.2102 \\
			E$^2$MetaNODE & 0.1230 & 0.2225 \\
			\hline \hline
			\multicolumn{3}{c}{\multirow{1}{*}{CUB-200-2011}} \\
			\hline
			SIB & 0.0920 & 0.1733 \\
			MetaNODE & 0.0920 & 0.2182 \\
			E$^2$MetaNODE & 0.0920 & 0.2220 \\
			\hline
	\end{tabular}}\smallskip
	\label{table4}
\end{table}

\noindent {\bf General Few-Shot Recognition.} Table~\ref{table1} shows the results of various evaluated methods on miniImagenet and tieredImagenet. In transductive FSL, we found that (\romannumeral1) MetaNODE and E$^2$MetaNODE all outperform the competitors (\emph{e.g.}, BD-CSPN, SIB, ProtoComNet, SRestoreNet) by around 2\% $\sim$ 7\%. This is because we rectify prototypes in a continuous-time prototype optimization manner instead of one-step rectification; (\romannumeral2) MetaNODE and E$^2$MetaNODE achieve superior performance over other state-of-the-art methods. Different from these methods, our method focuses on polishing prototypes instead of label propagation or loss evaluation. These results verify the superiority of our MetaNODE and E$^2$MetaNODE; (\romannumeral3) Our E$^2$MetaNODE achieves superior or comparable FSL performance over our MetaNODE conference version. Such performance improvement stems from the careful design from theory and the lower integration error of our meta-learning-based ODE solver. (\romannumeral4) The performance improvement is more conspicuous on 1-shot than 5-shot tasks, which is reasonable because the prototype bias is more remarkable on 1-shot tasks than 5-shot tasks because the labeled samples are more scarce.

\begin{table}[t]
	\centering
	\caption{Analysis of meta-optimizer on miniImagenet. 
		QS denotes unlabeled (query) samples from query set $Q$. 
	}
	\smallskip\scalebox
	{1.0}{
		\smallskip\begin{tabular}{c|l|c|c}
			\hline
			& Method &  5-way 1-shot & 5-way 5-shot \\
			\hline \hline
			\multicolumn{4}{c}{\multirow{1}{*}{MiniImagenet}} \\
			\hline
			(\romannumeral1) & Baseline & 61.22 $\pm$ 0.84$\%$ & $78.72 \pm 0.60\%$  \\
			(\romannumeral2) & + MetaLSTM & 63.85 $\pm$ 0.81$\%$ & 79.49 $\pm$ 0.65$\%$ \\
			(\romannumeral3) & + ALFA & 64.37 $\pm$ 0.79$\%$ & $80.75 \pm 0.57\%$ \\
			(\romannumeral4) & + MetaNODE & 66.07 $\pm$ 0.79$\%$ & 81.93 $\pm$ 0.55$\%$ \\
			(\romannumeral5) & + MetaNODE + QS & 77.92 $\pm$ 0.99$\%$ & $85.13 \pm 0.62\%$ \\
   			(\romannumeral6) & + E$^2$MetaNODE & 66.37 $\pm$ 0.80$\%$ & 81.97 $\pm$ 0.56$\%$ \\
			(\romannumeral7) & + E$^2$MetaNODE + QS & 78.62 $\pm$ 1.00$\%$ & $85.45 \pm 0.56\%$ \\
			\hline \hline
			\multicolumn{4}{c}{\multirow{1}{*}{TieredImagenet}} \\
			\hline
			(\romannumeral1) & Baseline & 69.71 $\pm$ 0.88$\%$ & $83.87 \pm 0.64\%$  \\
			(\romannumeral2) & + MetaLSTM & 70.56 $\pm$ 0.98$\%$ & 84.65 $\pm$ 0.73$\%$ \\
			(\romannumeral3) & + ALFA & 71.49 $\pm$ 0.93$\%$ & $85.58 \pm 0.60\%$ \\
			(\romannumeral4) & + Neural ODE & 72.72 $\pm$ 0.90$\%$ & 86.45 $\pm$ 0.62$\%$ \\
			(\romannumeral5) & + Neural ODE + QS & 83.46 $\pm$ 0.92$\%$ & $88.46 \pm 0.57\%$ \\
      		(\romannumeral6) & + E$^2$MetaNODE & 73.16 $\pm$ 0.92$\%$ & 86.66 $\pm$ 0.62$\%$ \\
			(\romannumeral7) & + E$^2$MetaNODE + QS & 84.10 $\pm$ 0.96$\%$ & $88.93 \pm 0.60\%$ \\
			\hline \hline
			\multicolumn{4}{c}{\multirow{1}{*}{CUB-200-2011}} \\
			\hline
			(\romannumeral1) & Baseline & 74.96 $\pm$ 0.86$\%$ & $88.89 \pm 0.43\%$  \\
			(\romannumeral2) & + MetaLSTM & 78.50 $\pm$ 0.75$\%$ &90.73 $\pm$ 0.41$\%$ \\
			(\romannumeral3) & + ALFA & 78.70 $\pm$ 0.77$\%$ & $90.98 \pm 0.41\%$ \\
			(\romannumeral4) & + Neural ODE & 80.82 $\pm$ 0.75$\%$ & 91.77 $\pm$ 0.49$\%$ \\
			(\romannumeral5) & + Neural ODE + QS & 90.94 $\pm$ 0.62$\%$ & $93.18 \pm 0.38\%$ \\
      		(\romannumeral6) & + E$^2$MetaNODE & 83.36 $\pm$ 0.71$\%$ & 92.96 $\pm$ 0.36$\%$ \\
			(\romannumeral7) & + E$^2$MetaNODE + QS & 91.27 $\pm$ 0.67$\%$ & $93.80 \pm 0.35\%$ \\
			\hline
	\end{tabular}}
	\smallskip
	\label{table5}
\end{table}
In inductive FSL, our MetaNODE and E$^2$MetaNODE method outperforms ClassifierBaseline by a large margin, around 3\% $\sim$ 5\%, on both datasets. This means that our method introducing a Neural ODE to polish prototype is effective. MetaNODE and E$^2$MetaNODE outperform our competitors (\emph{i.e.}, RestoreNet) by around 7\%, which validates the superiority of our manner to rectify prototypes. Besides, MetaNODE and E$^2$MetaNODE achieve competitive performance over other state-of-the-art methods. Here, (\romannumeral1) different from the metric and optimization methods, our method employs a pre-trained feature extractor and focuses on polishing prototype; (\romannumeral2) compared with other pre-training methods, our method focuses on obtaining reliable prototypes instead of fine-tuning feature extractors for FSL. The result validates the superiority of our methods. Particularly, our method beats RestoreNet, which also rectifies the prototypes. Different from the RestoreNet method, our method rectifies prototypes in a continuous manner instead of a one-step manner. Different from the RestoreNet method, our method rectifies prototypes in a continuous manner instead of a one-step manner. Compared with latest ProtoDiff \cite{du2023protodiff} (a concurrent work), our methods achieve comparable performance, which verifies the superiority of our methods. Finally,  our E$^2$MetaNODE performs better than MetaNODE version, which further verifies the effectiveness of our extension version E$^2$MetaNODE. %

\noindent {\bf Fine-Grained Few-Shot Image Recognition.} The results on CUB-200-2011 are shown in Table~\ref{table2}. Similar to Table~\ref{table1}, we observe that 1) our MetaNODE and E$^2$MetaNODE significantly outperform the state-of-the-art methods, achieving 2\% $\sim$ 6\% higher accuracy scores. This further verifies the effectiveness of MetaNODE and E$^2$MetaNODE in the fine-grained FSL task, which exhibits smaller class differences than the general FSL task; and 2) our E$^2$MetaNODE achieves better or comparable performance over MetaNODE, which indicates that our E$^2$MetaNODE is effective.

\begin{table}[t]
	\centering
	\caption{Analysis of GradNet components (i.e., ensemble, attention, and exponential decay) on miniImagenet, tieredImagenet, and CUB-200-2011.}
	\smallskip\scalebox
	{1.0}{
		\smallskip\begin{tabular}{c|c|c|c}
			\hline
			& Method & 5-way 1-shot & 5-way 5-shot \\
			\hline \hline
			\multicolumn{4}{c}{\multirow{1}{*}{MiniImagenet}} \\
			\hline
			(\romannumeral1) & MetaNODE & 77.92 $\pm$ 0.99$\%$ & 85.13 $\pm$ 0.62$\%$ \\
			(\romannumeral2) & w/o ensemble  & 75.34 $\pm$ 1.10$\%$ & 84.00 $\pm$ 0.53$\%$ \\
			(\romannumeral3) & w/o attention  & 75.10 $\pm$ 0.98$\%$ & 83.90 $\pm$ 0.56$\%$ \\
			(\romannumeral4) & w/o exponential decay  & 76.02 $\pm$ 1.17$\%$ & 84.16 $\pm$ 0.64$\%$ \\
			\hline \hline
			\multicolumn{4}{c}{\multirow{1}{*}{TieredImagenet}} \\
			\hline
			(\romannumeral1) & MetaNODE & 83.46 $\pm$ 0.92$\%$ & 88.46 $\pm$ 0.57$\%$ \\
			(\romannumeral2) &  w/o ensemble & 81.71 $\pm$ 0.93$\%$ & 86.42 $\pm$ 0.56$\%$ \\
			(\romannumeral3) & w/o attention & 81.47 $\pm$ 0.92$\%$ & 86.14 $\pm$ 0.61$\%$ \\
			(\romannumeral4) & w/o exponential decay & 81.27 $\pm$ 1.12$\%$ & 85.87 $\pm$ 0.74$\%$ \\
			\hline \hline
			\multicolumn{4}{c}{\multirow{1}{*}{CUB-200-2011}} \\
			\hline
			(\romannumeral1) & MetaNODE & 90.94 $\pm$ 0.62$\%$ & 93.18 $\pm$ 0.38$\%$ \\
			(\romannumeral2) & w/o ensemble & 89.91 $\pm$ 0.71$\%$ & 92.55 $\pm$ 0.38$\%$ \\
			(\romannumeral3) & w/o attention & 89.04 $\pm$ 0.80$\%$ & 92.28 $\pm$ 0.40$\%$ \\
			(\romannumeral4) & w/o exponential decay & 88.35 $\pm$ 0.89$\%$ & 92.18 $\pm$ 0.42$\%$ \\
			\hline
	\end{tabular}}
	\smallskip
	\label{table7}
\end{table}

\begin{table}[t]
	\centering
	\caption{Ablation study of GradNet components (i.e. gradient estimator, weight generator, and modules ensemble mechanism) on miniImagenet, tieredImagenet, and CUB-200-2011.}
	\smallskip\scalebox
	{0.86}{
		\smallskip\begin{tabular}{c|c|c|c}
			\hline
			& Method & 5-way 1-shot & 5-way 5-shot \\
			\hline \hline
			\multicolumn{4}{c}{\multirow{1}{*}{MiniImagenet}} \\
			\hline
			(\romannumeral1) & Baseline & 61.31 $\pm$ 0.85$\%$ & 78.71 $\pm$ 0.60$\%$ \\
			(\romannumeral2) & + Gradient Estimator (Eq.(9))  & 64.75 $\pm$ 0.79$\%$ & 80.90 $\pm$ 0.54$\%$ \\
			(\romannumeral3) & + Weight Generator (Eq.(10))  & 65.11 $\pm$ 0.77$\%$ & 81.44 $\pm$ 0.55$\%$ \\
            (\romannumeral4) & + Ensemble (w/o decay in Eq.(12))  & 65.76 $\pm$ 0.79$\%$ & 81.55 $\pm$ 0.55$\%$ \\
			(\romannumeral5) & + Ensemble (Eq.(12))  & 66.07 $\pm$ 0.79$\%$ & 81.93 $\pm$ 0.55$\%$ \\
			\hline \hline
			\multicolumn{4}{c}{\multirow{1}{*}{TieredImagenet}} \\
			\hline
			(\romannumeral1) & Baseline & 69.81 $\pm$ 0.88$\%$ & 83.71 $\pm$ 0.64$\%$ \\
			(\romannumeral2) & + Gradient Estimator (Eq.(9))  & 71.44 $\pm$ 0.87$\%$ & 85.49 $\pm$ 0.63$\%$ \\
			(\romannumeral3) & + Weight Generator (Eq.(10))  & 72.27 $\pm$ 0.92$\%$ & 86.28 $\pm$ 0.57$\%$ \\
            (\romannumeral4) & + Ensemble (w/o decay in Eq.(12))  & 72.60 $\pm$ 0.91$\%$ & 86.41 $\pm$ 0.60$\%$ \\
			(\romannumeral5) & + Ensemble (Eq.(12))  & 72.72 $\pm$ 0.90$\%$ & 86.45 $\pm$ 0.62$\%$ \\
			\hline \hline
			\multicolumn{4}{c}{\multirow{1}{*}{CUB-200-2011}} \\
			\hline
			(\romannumeral1) & Baseline & 74.81 $\pm$ 0.87$\%$ & 88.92 $\pm$ 0.43$\%$ \\
			(\romannumeral2) & + Gradient Estimator (Eq.(9))  & 77.53 $\pm$ 0.83$\%$ & 90.24 $\pm$ 0.40$\%$ \\
			(\romannumeral3) & + Weight Generator (Eq.(10))  & 78.69 $\pm$ 0.78$\%$ & 91.08 $\pm$ 0.40$\%$ \\
            (\romannumeral4) & + Ensemble (w/o decay in Eq.(12))  & 79.16 $\pm$ 0.76$\%$ & 91.45 $\pm$ 0.45$\%$ \\
			(\romannumeral5) & + Ensemble (Eq.(12))  & 80.82 $\pm$ 0.75 $\%$ & 91.77 $\pm$ 0.49$\%$ \\
			\hline
	\end{tabular}}
	\smallskip
	\label{table71}
\end{table}

\subsection{Statistical Analysis}
\label{section_4_4}
\noindent {\bf Does MetaNODE obtain more accurate prototypes?} In Table~\ref{table3}, we report the cosine similarity between initial (optimal) prototypes, \emph{i.e.}, $p(0)$ ($p(M)$) and real prototypes on 5-way 1-shot tasks of on miniImagenet, TieredImagenet, and CUB-200-2011 datasets. 
The reported results are averaged on 1000 episodes. 
We select BD-CSPN, SRestoreNet, and ProtoComNet as baselines, which also attempt to reduce prototype bias. The results show that 1) our MetaNODE and E$^2$MetaNODE obtain more accurate prototypes, which is because MetaNODE and E$^2$MetaNODE regard it as an optimization problem, and solves it in a continuous dynamics-based manner; and 2) our E$^2$MetaNODE achieves more high or comparable accuracy with MetaNODE on prototype estimation, which verifies E$^2$MetaNODE effectiveness.

\noindent {\bf Does MetaNODE alleviate gradient bias?} In Table~\ref{table4}, we randomly select 1000 episodes from miniImagenet, TieredImagenet, and CUB-200-2011 datasets, and then calculate the cosine similarity between averaged (inferred) and real gradient. Here, the averaged and infered gradients are obtained by Eq.~\ref{eq2} and GradNet, respectively. The real gradients are obtained in Eq.~\ref{eq2} by using all samples. We select SIB as the baseline, which improves GDA by inferring the loss value of unlabeled samples. It can be observed that 1) MetaNODE and E$^2$MetaNODE obtain more accurate gradients than SIB. This is because we model and train a meta-learner from abundant FSL tasks to directly estimate the continuous gradient flows; and 2) our E$^2$MetaNODE performs more superior, which shows its superiority. 

\begin{table}[t]
	\caption{Analysis of ODE solvers on miniImagenet, tieredImagenet, and CUB-200-2011.}\smallskip
	\centering
	\setlength{\tabcolsep}{3mm}
	\smallskip\scalebox
	{1.0}{
		\smallskip\begin{tabular}{c|c|c}
			\hline
			\multicolumn{3}{c}{\multirow{1}{*}{miniImageNet}} \\
			\hline
			ODE Solver & 5-way 1-shot & 5-way 5-shot \\
			\hline \hline
			Euler & 76.29 $\pm$ 1.03$\%$ & $84.00 \pm 0.56\%$ \\
			RK4 & 77.92 $\pm$ 0.99$\%$ & 85.13 $\pm$ 0.62$\%$ \\
                E$^2$Solver (Ours) & 78.19 $\pm$ 0.99$\%$ & 85.41 $\pm$ 0.57$\%$ \\
			\hline \hline
			\multicolumn{3}{c}{\multirow{1}{*}{tieredImageNet}} \\
			\hline
			ODE Solver & 5-way 1-shot & 5-way 5-shot \\
			\hline \hline
			Euler & 81.59 $\pm$ 0.97$\%$ & $87.04 \pm 0.58\%$ \\
			RK4 & 83.46 $\pm$ 0.92$\%$ & 88.46 $\pm$ 0.57$\%$ \\
                E$^2$Solver (Ours) & 84.45 $\pm$ 0.95$\%$ & 88.69 $\pm$ 0.69$\%$ \\
			\hline \hline
			\multicolumn{3}{c}{\multirow{1}{*}{CUB-200-2011}} \\
			\hline
			ODE Solver & 5-way 1-shot & 5-way 5-shot \\
			\hline \hline
			Euler & 87.71 $\pm$ 0.86$\%$ & $91.19 \pm 0.38\%$ \\
			RK4 & 90.94 $\pm$ 0.62$\%$ & 93.18 $\pm$ 0.38$\%$ \\
                E$^2$Solver (Ours) & 91.82 $\pm$ 0.62$\%$ & 94.23 $\pm$ 0.31$\%$ \\
			\hline
	\end{tabular}}
	\label{table88}
\end{table}

\begin{table}[t]
	\caption{Ablation study of our MetaNODE and E$^2$MetaNODE on miniImagenet, tieredImagenet and CUB-200-2011.}\smallskip
	\centering
	\smallskip\scalebox
	{0.9}{\begin{tabular}{c|c|c|c}
			\hline
			& Setting & 5-way 1-shot & 5-way 5-shot\\
			\hline \hline
			\multicolumn{4}{c}{\multirow{1}{*}{miniImagenet}} \\
			\hline
			(\romannumeral1) & Baseline & 61.22 $\pm$ 0.84$\%$ & 78.72 $\pm$ 0.60$\%$ \\
			(\romannumeral2) & GradNet + RK4 & 77.92 $\pm$ 0.99$\%$ & 85.13 $\pm$ 0.62$\%$ \\
			(\romannumeral3) & GradNet + E$^2$Solver & 78.19 $\pm$ 0.99$\%$ & 85.41 $\pm$ 0.57$\%$ \\
			(\romannumeral4) & E$^2$GradNet + RK4 & 78.36 $\pm$ 0.98$\%$ & 85.17 $\pm$ 0.50$\%$\\
			(\romannumeral5) & E$^2$GradNet + E$^2$Solver & 78.62 $\pm$ 1.00$\%$ & 85.45 $\pm$ 0.56$\%$ \\
			\hline \hline
			\multicolumn{4}{c}{\multirow{1}{*}{tieredImagenet}} \\
			\hline
			(\romannumeral1) & Baseline & 69.71 $\pm$ 0.88$\%$ & 83.87 $\pm$ 0.64$\%$ \\
			(\romannumeral2) & GradNet + RK4 & 83.46 $\pm$ 0.92$\%$ & 88.46 $\pm$ 0.57$\%$ \\
			(\romannumeral3) & GradNet + E$^2$Solver & 84.45 $\pm$ 0.95$\%$ & 88.69 $\pm$ 0.69$\%$ \\
			(\romannumeral4) & E$^2$GradNet + RK4 & 83.59 $\pm$ 0.96$\%$ & 88.53 $\pm$ 0.62$\%$\\
			(\romannumeral5) & E$^2$GradNet + E$^2$Solver & 84.10 $\pm$ 0.96$\%$ & 88.93 $\pm$ 0.60$\%$ \\
			\hline \hline
			\multicolumn{4}{c}{\multirow{1}{*}{CUB-200-2011}} \\
			\hline
                (\romannumeral1) & Baseline &74.96 $\pm$ 0.86$\%$ & 88.89 $\pm$ 0.43$\%$ \\
			(\romannumeral2) & GradNet + RK4 & 90.94 $\pm$ 0.62$\%$ & 93.18 $\pm$ 0.38$\%$ \\
			(\romannumeral3) & GradNet + E$^2$Solver & 91.82 $\pm$ 0.62$\%$ & 94.23 $\pm$ 0.31$\%$ \\
			(\romannumeral4) & E$^2$GradNet + RK4 & 91.15 $\pm$ 0.62$\%$ & 93.71 $\pm$ 0.35$\%$\\
			(\romannumeral5) & E$^2$GradNet + E$^2$Solver & 91.27 $\pm$ 0.67$\%$ & 93.96 $\pm$ 0.36$\%$ \\
			\hline
	\end{tabular}}
	\label{table777}
\end{table}

\subsection{Ablation Study}
\label{section_4_5}

\noindent {\bf Is our meta-optimizer effective?} In Table~\ref{table5}, we analyzed the effectiveness of our meta-optimizer. Specifically, in inductive FSL setting, (\romannumeral1) we remove our meta-optimizer as a baseline; 
(\romannumeral2) we add the MetaLSTM meta-optimizer \cite{RaviL17} on (\romannumeral1) to optimize prototypes; (\romannumeral3) we replace MetaLSTM by the ALFA \cite{BaikCCKL20} on (\romannumeral2); (\romannumeral4) different from (\romannumeral3), we replace MetaLSTM by our MetaNODE; and (\romannumeral5) we further explore unlabeled samples on (\romannumeral4); (\romannumeral6) different from (\romannumeral4), we replace MetaLSTM by our E$^2$MetaNODE; and (\romannumeral7) we further explore unlabeled samples on (\romannumeral6). 

\begin{table*}
    \caption{Computational efficiency analysis on miniImagenet. In ``a/b'', ``a'' and ``b'' denotes GPU memory and running time, respectively. ``BS'' denotes batch size during training. Note that the default BS=8 when ``BS'' is missing. }\smallskip
	\centering
	\smallskip\scalebox
	{0.93}{
		\smallskip\begin{tabular}{c|c|c|c|c|c|c}
                \hline
			\multirow{2}{*}{Setting} & \multirow{2}{*}{ClassifierBaseline \cite{Yinbo20}} & \multirow{2}{*}{ALFA \cite{BaikCCKL20}} & \multirow{2}{*}{CloserLook \cite{ChenLKWH19}} & \multicolumn{2}{c|}{\multirow{1}{*}{Our MetaNODE}}  & \multicolumn{1}{c}{\multirow{2}{*}{Our E$^2$MetaNODE}} \\
			\cline{5-6}
			& & & & Euler Solver & RK4 Solver (BS=2) &  \\
   			\hline
                \multicolumn{7}{c}{\multirow{1}{*}{Meta-Training Phase}} \\
			\hline
			5-way 1-shot & 11967 M/98.33 min & 7227 M/50.00 min & 11967 M/98.33 min & 34175 M/174.16 min & 32471 M/190.00 min  & 7203 M/65.83 min \\
			5-way 5-shot & 11967 M/98.33 min & 8577 M/55.00 min & 11967 M/98.33 min & 46493 M/221.67 min & 42685 M/229.17 min & 8553 M/79.16 min\\
			\hline
			\multicolumn{7}{c}{\multirow{1}{*}{Meta-Test Phase}}\\
			\hline
			5-way 1-shot & 1081 M/0.046 s & 1161 M/0.081 s & 1081 M/0.133 s & 1953 M/0.434 s & 1953 M/1.290 s & 1335 M/0.097s \\
			5-way 5-shot & 1181 M/0.059s & 1227M/0.085s & 1181M/0.359s & 2109M/0.452s & 2109M/1.300s & 1427M/ 0.113s\\
			\hline
	\end{tabular}}
	\label{table221}
\end{table*} 

From the results of (\romannumeral1) $\sim$ (\romannumeral4), we observe that: 1) the performance of (\romannumeral2) and (\romannumeral3) exceeds (\romannumeral1) around 1\% $\sim$ 3\%, which means that it is helpful to leverage the existing meta-optimizer to polish prototypes and validates the effectiveness of the proposed framework. 2) the performance of (\romannumeral4) exceeds (\romannumeral2) $\sim$ (\romannumeral3) around 2\% $\sim$ 3\%, which shows the superiority of our MetaNODE meta-optimizer. This is because MetaNODE regards the gradient flow as meta-knowledge, instead of hyperparameters like weight decay. 3) the performance of (\romannumeral6)/(\romannumeral7) exceeds (\romannumeral4)/(\romannumeral5) $\sim$ (\romannumeral3) around 2\% $\sim$ 3\%, which shows the superiority of our E$^2$MetaNODE meta-optimizer. This is because E$^2$MetaNODE is designed from theoretical perspective and employs a more higher-order meta-learning-based ODE solver. 4) comparing the results of (\romannumeral4)/(\romannumeral6) with (\romannumeral5)/(\romannumeral4), we find that using unlabeled samples can significantly enhance MetaNODE. 

\noindent {\bf Are these key components (\emph{i.e.}, ensemble, attention, and exponential decay) effective in GradNet of MetaNODE?} In Table~\ref{table7}, we evaluate the effect of these three components. Specifically, (\romannumeral1) we evaluate MetaNODE on miniImagenet; (\romannumeral2) we remove ensemble mechanism on (\romannumeral1), \emph{i.e.}, $H=1$; (\romannumeral3) we remove attention mechanism on (\romannumeral1); (\romannumeral3) we remove exponential decay mechanism on (\romannumeral1). We find that the performance decreases by around 1\% $\sim$ 3\% when removing these three components, respectively. This result implies that employing these three key components is beneficial for our MetaNODE. We note that our model only using a single module still achieves superior performance on 1-shot tasks. 

\noindent {\bf Are these key components (\emph{i.e.}, gradient estimator, weight generator, and modules ensemble mechanism) effective in GradNet?} In Table~\ref{table71},  we evaluate the effect of these four components, i.e., gradient estimator, weight generator, modules ensemble mechanism and its exponential decay. Specifically, (\romannumeral1) we first regard the simplistic gradient optimization as a baseline, i.e., optimizing the prototype by using a simple gradient descent manner; (\romannumeral2) we replace the simple gradient descent algorithm  of (\romannumeral1) by our MetaNODE to optimize prototypes, where we remove its weight generator and multi-modules ensemble mechanism and only gradient estimator is remained; (\romannumeral3) we add weight generator on (\romannumeral2); (\romannumeral4) we add multi-modules ensemble mechanism on (\romannumeral3), where we remove the exponential decay; and (\romannumeral5) we add the exponential decay on (\romannumeral4). The experimental results are reported in a new table (i.e., Table 8). From results, we can see that 1) the performance of (\romannumeral2) exceeds (\romannumeral1) around 1\%$\sim$3\%, which verifies the effectiveness of our gradient estimator component; 2) compared with (\romannumeral2)/(\romannumeral3)/(\romannumeral4), the performance of (\romannumeral3)/(\romannumeral4)/(\romannumeral5) all achieves performance improvement by around 1\% $\sim$ 2\%. This suggests that the proposed gradient estimation, weight generator, multi-modules ensemble mechanism, and exponential decay are all effective.

\noindent {\bf How do different solvers affect MetaNODE?} In Table~\ref{table88}, we show the classification performance of leveraging different ODE solvers to obtain the optimal prototypes. Specifically, we compare two common ODE solvers, i.e., Euler and Runge-Kutta (RK4) based solvers, which belong to lower-order and high-order integral methods, respectively. Theoretically, the RK4 method has lower integral error than Euler method. As shown in Table~\ref{table88}, (1) MetaNODE is more superior when RK4 is applied than Euler is used. This means that using high order solver is beneficial to our MetaNODE. Thus, we adopt the RK4 solver as our ODE solver in our MetaNODE; (2) MetaNODE achieves best performance when E$^2$Solver method is used. This verifies the effectiveness of our E$^2$Solver.

\noindent {\bf Are the proposed components (\emph{i.e.}, GradNet, E$^2$GradNet, and E$^2$Solver) effective in our MetaNODE and E$^2$MetaNODE?} In Table~\ref{table777}, we evaluate our GradNet, E$^2$GradNet and ODE solver components from conference and extension version, respectively. Specifically, (\romannumeral1) we remove our meta-optimizer as a baseline on miniImagenet, tieredImagenet, and CUB-200-2011; (\romannumeral2) we add our GradNet and RK4 solver used in conference version on (\romannumeral1); (\romannumeral3) we replace the ODE solver by our E$^2$Solver on (\romannumeral2); (\romannumeral4) we replace the GradNet by our E$^2$GradNet on (\romannumeral2); (\romannumeral5) we add our E$^2$GradNet and E$^2$Solver on (\romannumeral1). From results, we find that (1) compared with (\romannumeral1) and (\romannumeral2) $\sim$ (\romannumeral5), the performance of baseline achieves significant improvement after using our meta-optimizer, which verifies the effectiveness of our meta-optimizer; (2) the performance of (\romannumeral3) exceeds (\romannumeral2) around 1\%, which means that our $E^2$Solver is effective; (3) the performance of (\romannumeral4) exceeds (\romannumeral4) around 1\%, which verifies the effectiveness of our $E^2$GradNet; (4) the performance of (\romannumeral5) is best and this shows the superiority of our $E^2$GradNet and $E^2$Solver, which exactly is our E$^2$MetaNODE.   

\begin{table*}[t]
    \caption{Experiment results on the CIFAR-FS and FC100 data sets. The best results are highlighted in bold. `\_' denotes the absent results in original paper. }\smallskip
	\centering
	\smallskip\scalebox
	{0.9}{\begin{tabular}{l|l|c|c|c|c|c|c}
			\hline
			\multicolumn{1}{l|}{\multirow{2}{*}{Setting}}&
			\multicolumn{1}{l|}{\multirow{2}{*}{Method}}&
			\multicolumn{1}{c|}{\multirow{2}{*}{Type}}& \multicolumn{1}{c|}{\multirow{2}{*}{Backbone}}&
			\multicolumn{2}{|c|}{CIFAR-FS} & \multicolumn{2}{|c}{FC100} \\ 
			\cline{5-8}
			& & & & 5-way 1-shot & 5-way 5-shot & 5-way 1-shot & 5-way 5-shot \\
			\hline \hline
						\multicolumn{1}{l|}{\multirow{12}{*}{Transductive}} & SRestoreNet \cite{XueW20} & Metric & ResNet18 &  $69.09 \pm 0.97\%$  & $-$ & $-$  & $-$ \\
			& DPGN\cite{YangLZZZL20} & Graph & ResNet12 &  $77.9 \pm 0.50\%$  & \textbf{90.2} $\pm$ \textbf{0.40}\% & $-$  & $-$ \\
			& MCGN\cite{Tang_2021_CVPR} & Graph & ConvNet256 & $76.45 \pm 0.99\%$  & $88.42 \pm 0.23\%$ & $-$  & $-$ \\
			& ECKPN \cite{Chen_2021_CVPR} & Graph & ResNet12 & $79.20 \pm 0.40\%$  & $91.00 \pm 0.50\%$ & $-$  & $-$ \\
			& TFT\cite{DhillonCRS20} & Pre-training & WRN-28-10 &  $76.58 \pm 0.68\%$  & $85.79 \pm 0.50\%$ & $43.16 \pm 0.59\%$  & $57.57 \pm 0.55\%$ \\ 
			& SIB\cite{HuMXSOLD20} & Pre-training & WRN-28-10 & $80.00 \pm 0.60\%$  & $85.30 \pm 0.40\%$ & $-$  & $-$ \\
			& ICI\cite{WangXLZF20} & Pre-training & ResNet12 & $62.25\%$  & $80.82\%$ & $-$  & $-$ \\			 
			\cline{2-8}
			& MetaNODE (CV) & Pre-training & ResNet12 & 84.83 $\pm$ 0.88$\%$ & 89.15 $\pm$ 0.58$\%$ & 52.33 $\pm$ 0.99$\%$ & 64.34 $\pm$ 0.77$\%$ \\
			& E$^2$MetaNODE (Ours) & Pre-training & ResNet12 & \textbf{86.07} $\pm$ \textbf{0.86}$\%$ & \textbf{89.86} $\pm$ \textbf{0.56}$\%$ & \textbf{52.38} $\pm$ \textbf{1.08}$\%$ & \textbf{64.41} $\pm$ \textbf{0.48}$\%$ \\
			
			\hline
			\multicolumn{1}{l|}{\multirow{9}{*}{Inductive}} 
			& RestoreNet \cite{XueW20} & Metric & ResNet18 & $66.87 \pm 0.94\%$  & $-$ & $-$  & $-$ \\ 
			& ConstellationNet \cite{xu2021attentional} & Metric & ResNet12 &  $75.40 \pm 0.20\%$  & $86.80 \pm 0.20\%$ & $43.80 \pm 0.20\%$  & $59.70 \pm 0.20\%$ \\ 
			& RAP-ProtoNet \cite{snell2017prototypical} & Metric & ResNet10 &  $73.00 \pm 0.71\%$  & $85.46 \pm 0.47\%$ & $-$  & $-$ \\ 
			& MAML \cite{FinnAL17} & Optimization & ResNet12& $64.33 \pm 0.48\%$  & $76.38 \pm 0.42\%$ & $37.92 \pm 0.48\%$  & $52.63 \pm 0.50\%$ \\
			& MetaOptNet\cite{lee2019meta} & Optimization & ResNet12&  $72.00 \pm 0.70\%$  & $84.20 \pm 0.50\%$ & $41.10 \pm 0.60\%$  & $55.50 \pm 0.60\%$ \\
			& ALFA \cite{BaikCCKL20} & Optimization & ResNet12 & $68.25 \pm 0.47\%$  & $82.95 \pm 0.38\%$ & $42.37 \pm 0.50\%$  & $55.23 \pm 0.50\%$ \\
			
			\cline{2-8}
			& MetaNODE (CV) & Pre-training & ResNet12 & 75.48 $\pm$ 0.84$\%$ & 87.22 $\pm$ 0.60$\%$ & 43.96 $\pm$ 0.79$\%$ & \textbf{61.59} $\pm$ \textbf{0.74}$\%$ \\
			& E$^2$MetaNODE (Ours) & Pre-training & ResNet12 & \textbf{75.96} $\pm$ \textbf{0.81}$\%$ & \textbf{87.74} $\pm$ \textbf{0.55}$\%$ & \textbf{44.44} $\pm$ \textbf{0.77}$\%$ & 61.53 $\pm$ 0.74$\%$ \\
			\hline
	\end{tabular}}
	\label{table1111}
\end{table*}

\noindent {\bf Can our E$^2$MetaNODE improve computation  efficiency?} In Table~\ref{table221}, we report the running GPU memory and time of these baselines and our MetaNODE and E$^2$MetaNODE during meta-training and meta-test phases, where we select related ClassifierBaseline, ALFA, and CloserLook as our baselines. Note that we only set batch size to 2 for our MetaNODE with RK4 Solver due to the limitation of GPU memory and others methods with episodic training strategy default to 8. From experimental results, we can find that (\romannumeral1) compared with our baselines, our MetaNODE indeed takes a lot of computation time and GPU cost during meta-training and meta-test phases especially on the RK4 Solver (it has taken 34175M and 46493M GPU memory only when batch size is set 2); (\romannumeral2) its computation  efficiency is significantly improved after applying our E$^2$MetaNODE. This result implies that employing our $E^2$GradNet and $E^2$Solver is beneficial for improving computation  efficiency of our MetaNODE.

\subsection{More Comparsion on CIFAR Derivative Datasets}

In this subsection, we conduct more experiments on CIFAR derivative datasets\cite{Yinbo20}, i.e., CIFAR-FS and FC100. The experimental results are shown in Table~\ref{table1111}. From these results, we can observe similar conclusions to Table~\ref{table1}, i.e., (i) significant performance improvement over state-ofthe-art methods (around 1\% $\sim$ 6\%); (iii) consistent performance improvement on transductive and inductive FSL. This further verifies the effectiveness and robuseness of our methods.

\subsection{Visualization}

\begin{figure}[h]
	\centering
	
	\subfigure[Performance (MetaNODE)]{ 
		\label{fig4a} 
		\includegraphics[width=0.47\columnwidth]{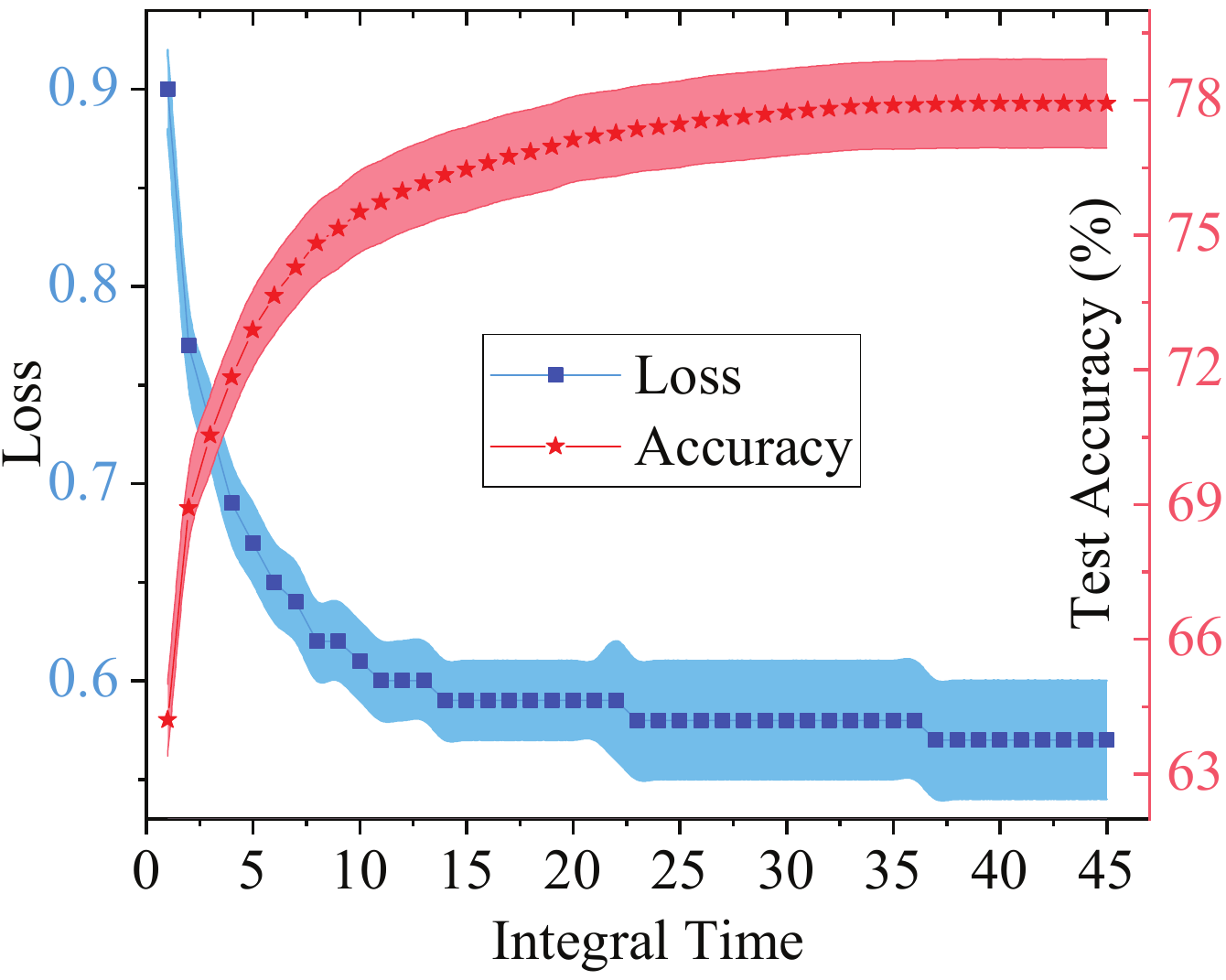}}
	\subfigure[Feature space (MetaNODE)]{ 
		\label{fig4b} 
		\includegraphics[width=0.47\columnwidth]{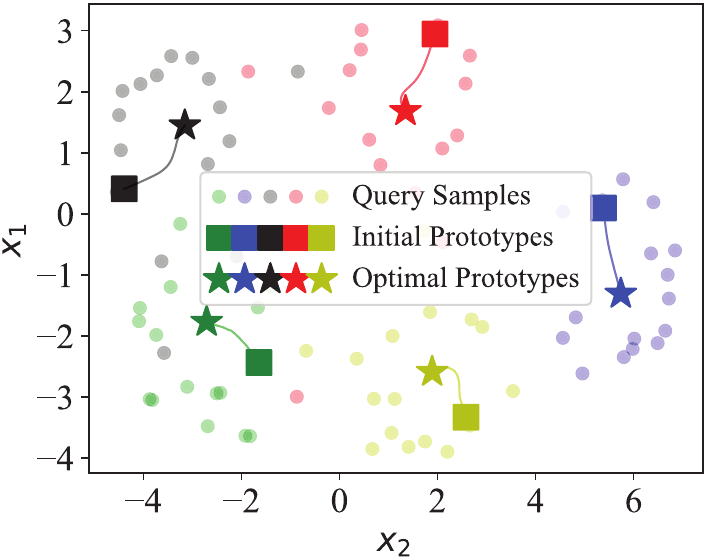}}
	\caption{Visualization of our MetaNODE on miniImagenet.}
	\label{fig4}
\end{figure}

\noindent {\bf Can our meta-optimizer converge?} In Figure~\ref{fig4a}, we randomly select 1000 episodes (5-way 1-shot) from miniImageNet, and then report their test accuracy and loss from integral time $t = $ 1 to 45. It can be observed that our meta-optimizer can converge to a stable result when $t=40$. Hence, $M=40$ is a default setting in our approach.  

\noindent {\bf How our meta-optimizer works?}
We visualize the prototype dynamics of a 5-way 1-shot task of miniImagenet, in Figure~\ref{fig4b}. Note that (1) since there is only one support sample in each class, its feature is used as the initial prototypes; (2) the curve from the square to the star denotes the trajectory of prototype dynamics and its tangent vector represents the gradient predicted by our GradNet. We find that the initial prototypes marked by squares flow to optimal prototypes marked by stars along prototype dynamics, much closer to the class centers. This indicates that our meta-optimizer effectively learns the prototype dynamics. 

\section{Conclusion}
\label{section_6}
In this paper, we propose a novel meta-learning based prototype optimization framework to obtain more accurate prototypes for few-shot learning. In particular, we design a novel Neural ODE-based meta-optimizer (called MetaNODE) to capture the continuous prototype dynamics. Further more, to improve the computational efficiency of our MetaNODE, we conduct a theoretical analysis on our meta-optimizer, and then present an effective and efficient extension version E$^2$MetaNODE. Specifically, we simplify and design a more efficient and efective gradient flow inference network (i.e., $E^2$GradNet) and propose a meta-learning-based ODE solver (i.e., $E^2$Solver) for prototype optimization. Experimental results on various datasets show that our methods (i.e., MetaNODE and E$^2$MetaNODE) obtain superior performance over previous state-of-the-art methods and our E$^2$MetaNODE significantly improves computation efficiency meanwhile without performance degradation. We also conduct extensive statistical experiments and ablation studies, which verify the superiority of our methods. 


%

\ifCLASSOPTIONcompsoc
  \section*{Acknowledgments}
\else
  \section*{Acknowledgment}
\fi

This work was supported by the National Natural Science
Foundation of China under Grant No. 62272130 and Grant No. 61972111, and the Shenzhen Science and Technology Program under Grant No. KCXFZ20211020163403005, Grant No. JCYJ20210324120208022 and Grant No. JCYJ20200109113014456.

\ifCLASSOPTIONcaptionsoff
  \newpage
\fi



\bibliographystyle{IEEEtran}
\bibliography{egbib}

\begin{IEEEbiography}[{\includegraphics[width=1in,height=1.25in,clip,keepaspectratio]{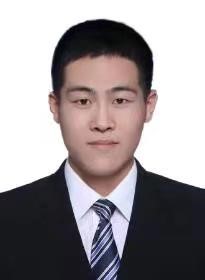}}]{Baoquan Zhang}
    is currently an Assistant Professor with Computer Science and Technology, Harbin Institute of Technology, Shenzhen, China. He received the Ph.D. in Computer Science and Technology from Harbin Institute of Technology, Shenzhen, China, in 2023, the M.S. degree from the Harbin Institute of Technology, China, in 2017, and the B.S. degree from the Harbin Institute of Technology, Weihai, China, in 2015. His current research interests include meta learning, few-shot learning, and data mining.
\end{IEEEbiography}

\begin{IEEEbiography}[{\includegraphics[width=1in,height=1.25in,clip,keepaspectratio]{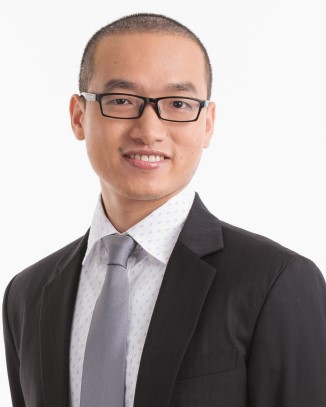}}]{Shanshan Feng}
    is currently a senior Research Scientist with A*STAR, Singapore. He received the Ph.D. degree in Computer Science from Nanyang Technological University, Singapore, in 2017. His research interests include machine learning, sequential data mining and social network analysis.
\end{IEEEbiography}

\begin{IEEEbiography}[{\includegraphics[width=1in,height=1.25in,clip,keepaspectratio]{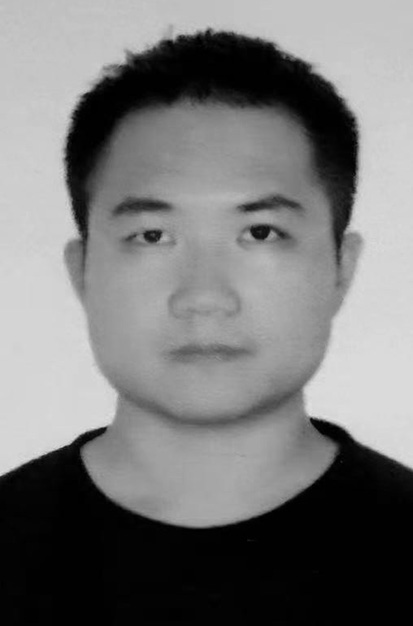}}]{Bingqi Shan}
    is currently pursuing a Master degrees with the School of Computer Science and Technology, Harbin Institute of Technology, Shenzhen, China. He received the B.S. degree from the Huaqiao University, China, in 2019. His current research interests include meta learning, few-shot learning.
\end{IEEEbiography}

\begin{IEEEbiography}[{\includegraphics[width=1in,height=1.25in,clip,keepaspectratio]{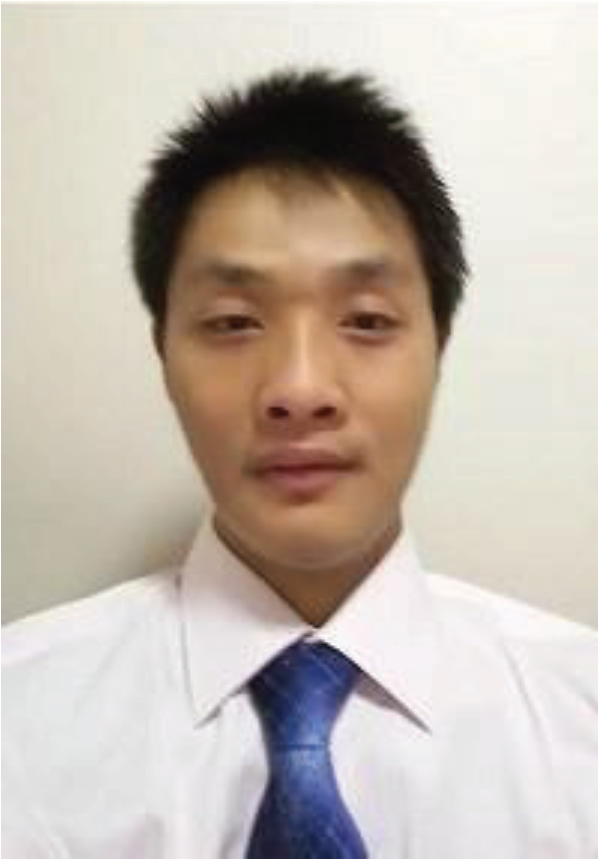}}]{Xutao Li}
	is currently a Professor with the School of Computer Science and Technology, Harbin Institute of Technology, Shenzhen, China. He received the Ph.D. and Master degrees in Computer Science from Harbin Institute of Technology in 2013 and 2009, and the Bachelor from Lanzhou University of Technology in 2007. His research interests include data mining, machine learning, graph mining, and social network analysis, especially tensor-based learning, and mining algorithms.
\end{IEEEbiography}

\begin{IEEEbiography}[{\includegraphics[width=1in,height=1.25in,clip,keepaspectratio]{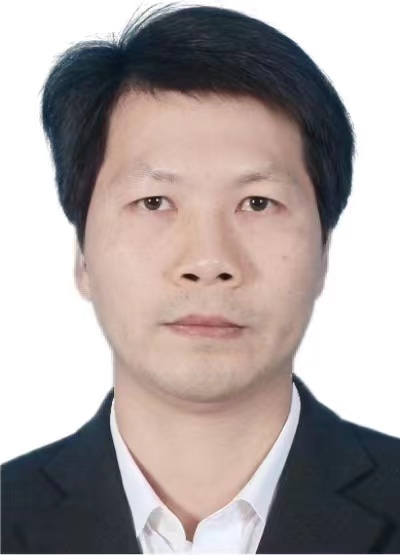}}]{Yunming Ye}
	is currently a Professor with the School of Computer Science and Technology, Harbin Institute of Technology, Shenzhen, China. He received the PhD degree in Computer Science from Shanghai Jiao Tong University, Shanghai, China, in 2004. His research interests include data mining, text mining, and ensemble learning algorithms.
\end{IEEEbiography}

\begin{IEEEbiography}[{\includegraphics[width=1in,height=1.25in,clip,keepaspectratio]{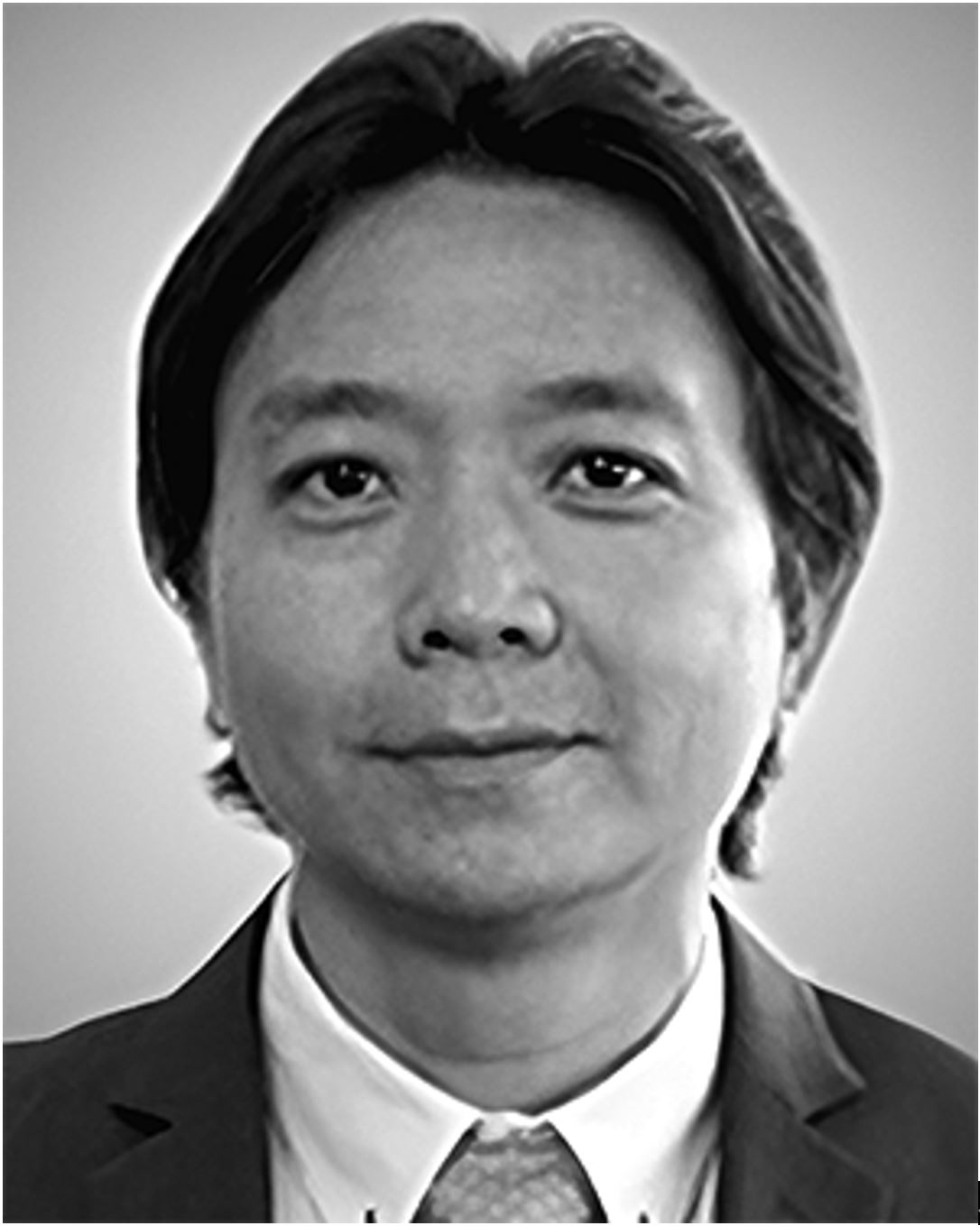}}]
{Yew-soon Ong}
(Fellow, IEEE) received the Ph.D. degree from the University of Southampton, U.K., in 2003. He is the president’s chair professor in Computer Science with Nanyang Technological University (NTU), and holds the position of chief artificial intelligence scientist of A*STAR, Singapore. At NTU, he serves as co-director of the Singtel-NTU Cognitive Artificial Intelligence Joint Lab. His research interest is in artificial and computational intelligence. He is the founding EIC of IEEE Transactions on Emerging Topics in Computational Intelligence and AE of IEEE Transactions on Neural Networks and Learning Systems, IEEE on Transactions on Cybernetics, IEEE Transactions on Artificial Intelligence and others. He has received several IEEE outstanding paper awards and was listed as a Thomson Reuters highly cited Researcher and among the World’s Most Influential Scientific Minds.
\end{IEEEbiography}

\end{document}